\documentclass[technote,onecolumn,a4paper,11pt]{IEEEtran}
\IEEEoverridecommandlockouts

\usepackage{amsmath,amssymb,amsfonts,bm}
\usepackage{algorithmic}
\usepackage{cite}
\usepackage{graphicx}
\usepackage{textcomp}
\usepackage{mathtools}
\usepackage{xcolor}
\usepackage{subcaption}
\usepackage{float}
\usepackage{balance}
\usepackage{hyperref}
\usepackage{booktabs}
\usepackage{multirow}
\usepackage{tabularx}
\usepackage{array} 
\usepackage[tableposition=top]{caption}
\def\BibTeX{{\rm B\kern-.05em{\sc i\kern-.025em b}\kern-.08em
    T\kern-.1667em\lower.7ex\hbox{E}\kern-.125emX}}

\begin{document}

\title{Parameter-Minimal Neural DE Solvers via Horner Polynomials}
 
\author{\IEEEauthorblockN{1\textsuperscript{st} Tomislav Matuli\'{c}}
\IEEEauthorblockA{\textit{Dept. of Electronic Systems and Information Processing} \\
\textit{Faculty of Electrical Engineering and Computing}\\
\textit{University of Zagreb}\\
Zagreb, Croatia \\
tomislav.matulic@fer.hr\\\vspace{10pt}}
\and
\IEEEauthorblockN{2\textsuperscript{nd} Damir Ser\v{s}i\'{c}}
\IEEEauthorblockA{\textit{Dept. of Electronic Systems and Information Processing} \\
\textit{Faculty of Electrical Engineering and Computing}\\
\textit{University of Zagreb}\\
Zagreb, Croatia \\
damir.sersic@fer.hr}
}

\maketitle

\begin{abstract}
We propose a parameter-minimal neural architecture for solving differential equations by restricting the hypothesis class to Horner-factorized polynomials, yielding an implicit, differentiable trial solution with only a small set of learnable coefficients. Initial conditions are enforced exactly by construction by fixing the low-order polynomial degrees of freedom, so training focuses solely on matching the differential-equation residual at collocation points. To reduce approximation error without abandoning the low-parameter regime, we introduce a piecewise (“spline-like”) extension that trains multiple small Horner models on subintervals while enforcing continuity (and first-derivative continuity) at segment boundaries. On illustrative ODE benchmarks and a heat-equation example, Horner networks with tens (or fewer) parameters accurately match the solution and its derivatives and outperform small MLP and sinusoidal-representation baselines under the same training settings, demonstrating a practical accuracy–parameter trade-off for resource-efficient scientific modeling.
\end{abstract}


\section{Introduction}

Neural networks have found widespread applications in many domains, including biology \cite{Jiang2022, Walters2020}, medicine\cite{Aggarwal2021, Shen2020, S2021AnalysisOD}, hyperspectral data analysis\cite{BioucasDias2013}, and even creative fields such as music, art, and digital media\cite{Briot2018}. Their ability to recognize patterns and make predictions from large and complex datasets has fundamentally transformed how we address many scientific, technological, and creative challenges. Large Language Models(LLMs), such as GPT\cite{OpenAI2025}, or DeepSeek\cite{deepseekai2024}, have seen unprecedented growth and innovation in recent years. These models, capable of comprehending and generating human-like language with extraordinary accuracy, are opening new frontiers in communication, education, research, and creative writing. 

Differential equations (DEs) are a central modeling tool in science and engineering, but many practical settings require solution strategies that are both accurate and lightweight (e.g., rapid prototyping, embedded deployment, or repeated solution under limited compute).
In recent years, scientific machine learning has popularized neural approaches that represent the unknown solution as a differentiable function approximator and train it by minimizing the governing-equation residual at collocation points.

Implicit Neural Representation (INR) is an approach that encodes data, such as images, audio signals, or 3D objects, using a neural network \cite{NEURIPS2022_575c4500, Strmpler2022a, Chen_2024_CVPR, Yang_2023_ICCV}. Here, the network itself acts as a continuous function, mapping the input coordinates to the corresponding data values. Unlike traditional explicit methods that store data as discrete arrays (e.g., pixel grids or voxel grids), INRs provide a compact, flexible, and resolution-independent way to represent and model complex signals. This approach has a wide range of applications, including 3D scene representation, high-resolution image reconstruction, and audio/video encoding. A particularly exciting application of INRs is in solving ordinary differential equations (ODEs) and partial differential equations (PDEs) by representing their solutions as continuous functions learned by neural networks. One common way to implement this is to use a Multilayer Perceptron (MLP), where the network learns the solution through coordinate mappings. Activation functions such as ReLU, Tanh, and Softplus are often used in these architectures to capture the underlying structures of solutions. In \cite{Jiang2020}, researchers have developed super-resolution frameworks to generate grid-free solutions to PDEs, demonstrating the power of neural networks in avoiding the need for traditional discretized grids. The use of periodic activation functions (such as sine and cosine) has been shown to offer significant advantages over traditional activation functions when representing high-frequency signals in applications such as audio, video, and 3D object modeling. These periodic functions also excel in solving complex differential equations by accurately capturing oscillatory and periodic behaviors that other activations struggle to represent.\cite{siren} Tensor Neural Networks, presented in \cite{https://doi.org/10.48550/arxiv.2208.02235}, have also been developed to tackle PDEs, offering powerful architectures capable of solving complex physical models. 

Physics-Informed Neural Networks \cite{Cuomo2022, Donnelly2024, Chen20241, Raissi2019} (PINNs) represent a powerful class of deep learning models designed to solve problems governed by physical laws, often described by ODE or PDE. Unlike conventional neural networks, which rely primarily on data-driven learning, PINNs embed known physical principles directly into the training process. This integration ensures that the network’s predictions adhere to the underlying laws of physics while learning from the observed data. Traditional neural networks train by minimizing the difference between predictions and ground truth data, but PINNs take this a step further by incorporating physics-based constraints directly into the loss function, creating more reliable and physically accurate models. A canonical example is the class of neural networks \cite{Raissi2019}, where a neural network  represents the unknown field and is trained by minimizing a loss that penalizes the differential operator residual at collocation points, optionally augmented by data misfit and initial/boundary condition terms. Several extensions address accuracy and training efficiency by enriching the constraint information. Gradient-enhanced PINN \cite{Yu2022aa} augment the standard PINN objective by also penalizing derivatives of the PDE residual; this can improve convergence and reduce the number of points required to achieve a given accuracy.

A distinct and increasingly influential paradigm is operator learning, where the goal is not to solve a single PDE instance, but to learn the mapping from an input function (e.g., initial condition, coefficient field, forcing) to the corresponding solution function. 
Deep Operator Networks \cite{Lu2021a} realize this by combining a branch network that encodes the input function sampled at sensors and a trunk network that encodes the query location, resulting in a mesh-independent surrogate that can generalize across families of PDE instances after amortized training. 
The Fourier Neural Operator \cite{JMLR:v24:21-1524} parameterizes an integral operator in Fourier space, enabling efficient learning of solution operators for parametric PDEs and providing strong accuracy--speed trade-offs, with the important practical feature that evaluation can be performed on resolutions different from those used during training.

A line of research explores structured function classes inside neural architectures, such as inductive bias, including polynomial- and spline-like parameterizations. 
Classical polynomial neural networks \cite{Oh2002} build multi-layer models from low-order polynomial units (often quadratic) with data-driven structure selection, while higher-order sigma--pi--sigma networks \cite{Li2003aa} explicitly introduce multiplicative interactions to represent polynomial terms more directly. 
More recently, deep polynomial networks (e.g. $\Pi$-nets \cite{poly2020}) revisit polynomial representations in modern deep learning by realizing high-order polynomials through architectural compositions and skip connections, and theoretical work \cite{NEURIPS2019_a0dc078c} studies the expressive power of deep polynomial networks through an algebraic lens. 
In a related spirit, Kolmogorov--Arnold Networks (KANs) \cite{liu2025kan} replace fixed node activations with learnable univariate functions on edges, typically parameterized by splines, offering an alternative to MLPs that is explicitly biased toward low-dimensional functional structure and has attracted attention for scientific regression tasks.

Inspired by the fact that biological systems can produce rich and robust behavior with a surprisingly small number of neurons; from the compact \emph{C.~elegans} connectome to small central pattern generators and sophisticated insect cognition; we hypothesize that strong structural priors and hard constraints can substitute for sheer parameter count.
Guided by this principle, we propose parameter-minimal neural architectures for solving differential equations that incorporate rules and constraints directly into the structure of the model rather than relying on delicate loss-term balancing.

Our contribution is architecture-first and parameter-minimal: instead of relying on a large, generic network whose compliance with the governing equation and constraints emerges only through a delicately balanced multi-term loss, we explicitly restrict the hypothesis class to a Horner-factorized polynomial representation and learn only a small set of coefficients. The most important constraints are enforced by construction: initial conditions are embedded directly into the model so that they are satisfied exactly, removing a major source of penalty-weight tuning and reducing the risk of partially feasible solutions that satisfy the residual but drift at the initial condition. Beyond efficiency, this structural restriction also improves interpretability: the learnable parameters correspond to identifiable polynomial (and piecewise-polynomial) degrees of freedom, enabling transparent control over model capacity. This architectural constraint viewpoint aligns with the broader theme of embedding physics/constraints into the network structure, but here it is paired with an unusually tight control of capacity. As a result, the learning problem becomes low-dimensional and strongly regularized, which empirically yields faster and more reliable convergence than over-parameterized baselines in the intended single-instance setting, while achieving high accuracy with orders-of-magnitude fewer learnable parameters.

In summery, the main contributions of this work are:
\begin{itemize}
\item \textbf{Architecture-first, parameter-minimal solver for ODEs and PDEs.}
We propose a Horner-factorized polynomial neural architecture for ODE/PDE solving that achieves a smooth, differentiable solution representation with a tightly limited number of learnable coefficients.

\item \textbf{Hard enforcement of initial conditions via architectural embedding.}
We embed initial conditions directly into the model so that they are satisfied exactly, reducing hyperparameter sensitivity (e.g., penalty weights) and improving training reliability.

\item \textbf{High-accuracy solutions (including derivatives) with far fewer parameters.}
The proposed model accurately captures both the solution and its derivatives, while using fewer learnable parameters.

\item \textbf{Interpretability and user control.}
Model capacity is explicitly controlled by polynomial order and/or the number of segments, yielding an interpretable parameterization with transparent accuracy--complexity trade-offs.

\item \textbf{Spline-like piecewise extension.}
We extend the Horner model to a piecewise Horner construction that improves approximation accuracy with minimal parameter growth, enforcing continuity and smoothness across subinterval boundaries.

\end{itemize}

Section II presents the notation used throughout this paper. Section III demonstrates solving differential equations using standard MLP neural networks (competitive methods), which serve as a baseline for comparison. Section IV introduces the motivation for our neural network architecture and paradigm, based on polynomial linear regression models. Section V presents a novel neural network architecture inspired by the Horner scheme. Section VI extends these models using a spline-like approach to further enhance their flexibility and accuracy. Finally, Section VII concludes the paper with a summary of our findings and potential future directions.

\section{Notation and Benchmark Problems}
This paper develops a parameter-minimal neural-network paradigm for solving ordinary differential equations (ODEs). We first fix the notation and summarize the benchmark problems used throughout the paper.

\subsection{Problem setup}
Let $x:I\rightarrow\mathbb{R}$ defined on interval $0\in I\subseteq\mathbb{R}$ denote the unknown solution and let $x^{(i)}(t)$ denote its $i$-th derivative.
We consider an $n$-th order ODE that can be written in the general form
\[
F\!\bigl(t, x(t), x'(t), \ldots, x^{(n)}(t)\bigr)
=
G\!\bigl(u(t), u'(t), \ldots, u^{(l)}(t)\bigr),
\]
where $u(t)$ is a known input signal. Since the right-hand side depends only on known quantities, we define the (known) forcing term
\[
f(t)\coloneqq G\!\bigl(u(t), u'(t), \ldots, u^{(l)}(t)\bigr),
\]
and hence study the explicit-forcing form
\begin{equation}
F\!\bigl(t, x(t), x'(t), \ldots, x^{(n)}(t)\bigr) = f(t).
\label{eq:ODEgeneral}
\end{equation}
To uniquely determine $x(t)$, we assume $n$ initial conditions specified at $t=0$:
\[
x^{(i)}(0) = x_i,\qquad i=0,1,\ldots,n-1.
\]

\subsection{Collocation data}
Residual-based training evaluates the known forcing at a set of collocation points
$\mathcal{T}=\{t_k\}_{k=1}^{M}\subset I$.
The available data are therefore the samples
\[
\mathcal{D}=\{(t_k,f_k)\}_{k=1}^{M},\qquad f_k \coloneqq f(t_k).
\]
Note that the sampling of $\{t_k\}$ need not be uniform.

\subsection{Benchmark problems}
To illustrate the proposed approach, we solve three representative ODEs: a first-order linear ODE, a first-order nonlinear ODE, and a second-order linear ODE.

Type A (first-order linear).
\begin{equation}
\begin{aligned}
x'(t) + 2x(t) &= 1,\\
x(0) &= 1,
\end{aligned}
\label{eq:benchA}
\end{equation}
with exact solution $x(t)=\tfrac{1}{2}\bigl(1+e^{-2t}\bigr)$. We consider this benchmark on the interval of interest $I=[0,4]$.

Type B (first-order nonlinear).
\begin{equation}
\begin{aligned}
x'(t)\,x(t) &= t,\\
x(0) &= 1,
\end{aligned}
\label{eq:benchB}
\end{equation}
whose solution satisfying the initial condition is the positive branch
$x(t)=\sqrt{t^{2}+1}$. We consider this benchmark on the interval of interest $I=[0,3]$.

Type C (second-order linear).
\begin{equation}
\begin{aligned}
x''(t) + 4x'(t) + 13x(t) &= 2,\\
x(0) &= 0,\\
x'(0) &= 1,
\end{aligned}
\label{eq:benchC}
\end{equation}
with exact solution
\[
x(t)=\frac{2}{13}+e^{-2t}\!\left(\frac{3}{13}\sin(3t)-\frac{2}{13}\cos(3t)\right).
\]
 We consider this benchmark on the interval of interest $I=[0,3]$.

\begin{table}[t]
\centering
\small
\caption{Benchmark ODEs used throughout the paper.}
\label{tab:bench_odes}
\setlength{\tabcolsep}{4pt}
\renewcommand{\arraystretch}{1.2}
\begin{tabularx}{\linewidth}{@{} c >{\raggedright\arraybackslash}X >{\raggedright\arraybackslash}X c >{\raggedright\arraybackslash}X @{}}
\toprule
\textbf{Type} & \textbf{ODE} & \textbf{Initial conditions} & \textbf{Exact solution} \\
\midrule
\textbf{A} &
$\displaystyle x'(t)+2x(t)=1$ &
$\displaystyle x(0)=1$ &
$\displaystyle x(t)=\tfrac{1}{2}\bigl(1+e^{-2t}\bigr)$
\\[2pt]

\textbf{B} &
$\displaystyle x'(t)\,x(t)=t$ &
$\displaystyle x(0)=1$ &
$\displaystyle x(t)=\sqrt{t^{2}+1}$
\\[2pt]

\textbf{C} &
$\displaystyle x''(t)+4x'(t)+13x(t)=2$ &
$\displaystyle x(0)=0,\;\;x'(0)=1$ &
$\displaystyle x(t)=\frac{2}{13}+e^{-2t}\!\left(\frac{3}{13}\sin(3t)-\frac{2}{13}\cos(3t)\right)$
\\
\bottomrule
\end{tabularx}
\end{table}

Table~\ref{tab:bench_odes} summarizes the benchmark ODEs used throughout the paper, including the governing equation, the initial conditions, and the corresponding closed-form solution. Unless stated otherwise, all problems are considered on the same interval of interest $I$, and all reported training and evaluation results in the subsequent sections refer to these benchmarks and the notation introduced above.

\section{Baseline Solvers: MLP and SIREN Networks}
To contextualize the proposed parameter-minimal Horner architectures, we first evaluate standard coordinate-based neural solvers that represent the unknown solution by a neural network $N(t)$ and minimize the differential-equation residual at collocation points. These baselines follow the common residual-minimization paradigm used in physics-informed training and implicit neural representations.

\subsection{Residual loss with soft initial-condition penalties}
Given collocation points $\mathcal{T}=\{t_i\}_{i=1}^{M}\subset I$, we define the residual
\[
r(t_i) \;=\; F\!\bigl(t_i,\,N(t_i),\,\tfrac{d}{dt}N(t_i),\,\ldots,\,\tfrac{d^n}{dt^n}N(t_i)\bigr)\;-\;f(t_i),
\]
and minimize the mean-squared residual augmented by soft penalties on the initial conditions:
\begin{equation}
\mathcal{L}_{\mathrm{base}}
=
\frac{1}{M}\sum_{i=1}^{M} r(t_i)^2
\;+\;
\sum_{j=0}^{n-1}\lambda_j\Bigl(\tfrac{d^j}{dt^j}N(0)-x_j\Bigr)^2 .
\label{eq:baseline_loss}
\end{equation}
Here, $\lambda_j\ge 0$ are hyperparameters controlling the strength of the initial-condition enforcement. In contrast, the Horner-based models introduced later embed these conditions by construction and therefore do not require such penalty terms.

\subsection{Baseline architectures and training protocol}
We report results for Type~A and compare three representative baselines:
(i) a wide MLP with a piecewise-linear activation (Leaky ReLU),
(ii) a compact MLP with a smooth activation (sigmoid),
and (iii) a compact sinusoidal representation network (SIREN).
All models are trained using Adam with an initial learning rate of $10^{-3}$ for $10{,}000$ epochs on $M=400$ collocation points. Unless stated otherwise, we set $\lambda_0=0.1$ and use the remaining $\lambda_j$ only when higher-order initial conditions are present.

\paragraph{Baseline 1 (MLP--Leaky ReLU).}
We use an MLP with 5 hidden layers of width 256 (263{,}937 learnable parameters) and Leaky ReLU activations.

\paragraph{Baseline 2 (MLP--sigmoid).}
We use an MLP with 4 hidden layers of width 5 (106 learnable parameters) and sigmoid activations.

\paragraph{Baseline 3 (SIREN).}
We use a SIREN with 4 hidden layers of width 5 (106 learnable parameters), following the standard sinusoidal-activation design and initialization.

\subsection{Results and discussion}
\begin{figure}[t]
    \centering
    \captionsetup{font=small}
    \captionsetup[subfigure]{font=footnotesize,justification=centering}

    \textbf{MLP (Leaky ReLU)}\\[-2pt]
    \vspace*{5pt}
    \begin{subfigure}[t]{0.32\linewidth}
        \centering
        \includegraphics[width=\linewidth]{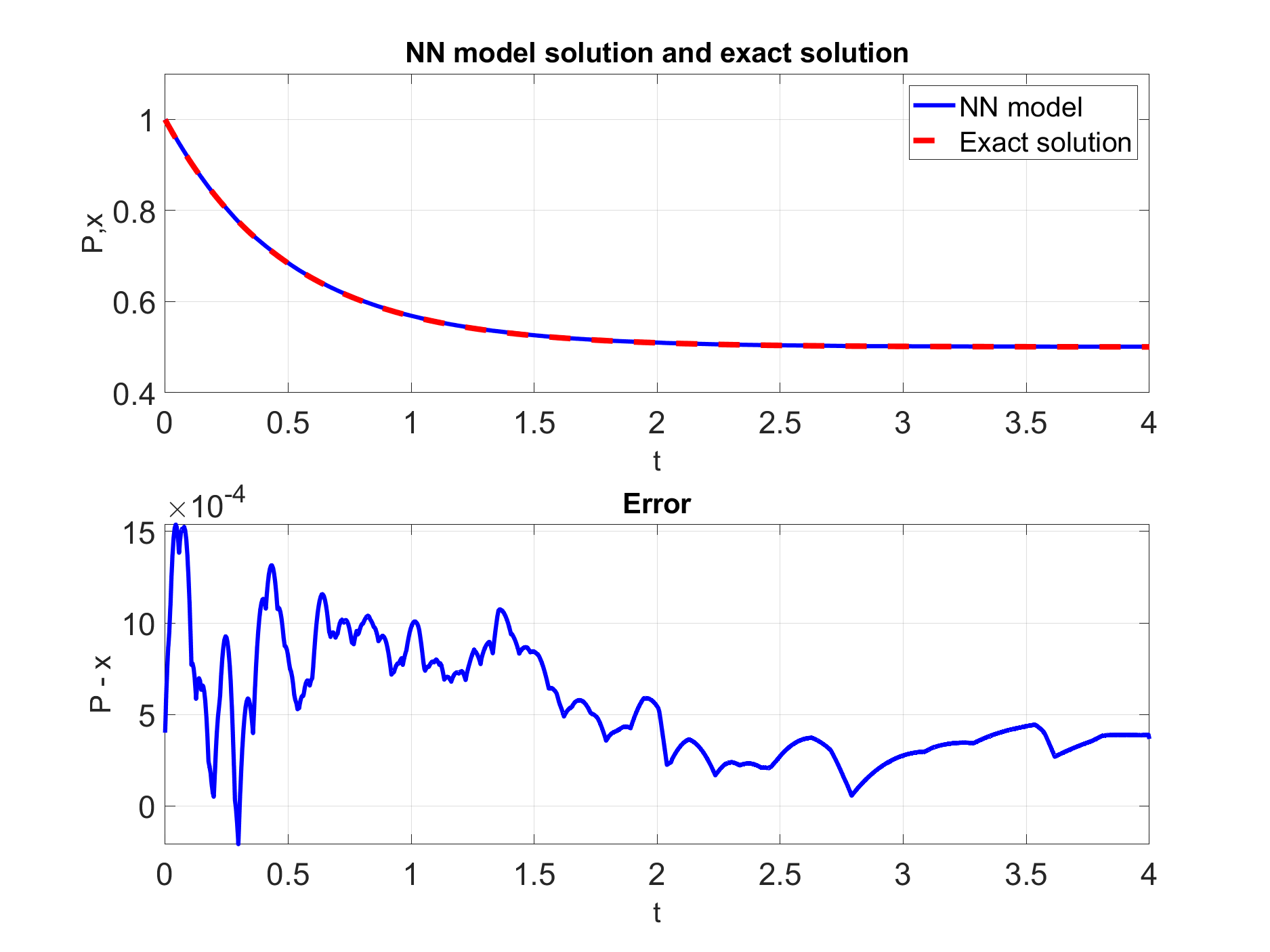}
        \caption{$x(t)$}
    \end{subfigure}\hfill
    \begin{subfigure}[t]{0.32\linewidth}
        \centering
        \includegraphics[width=\linewidth]{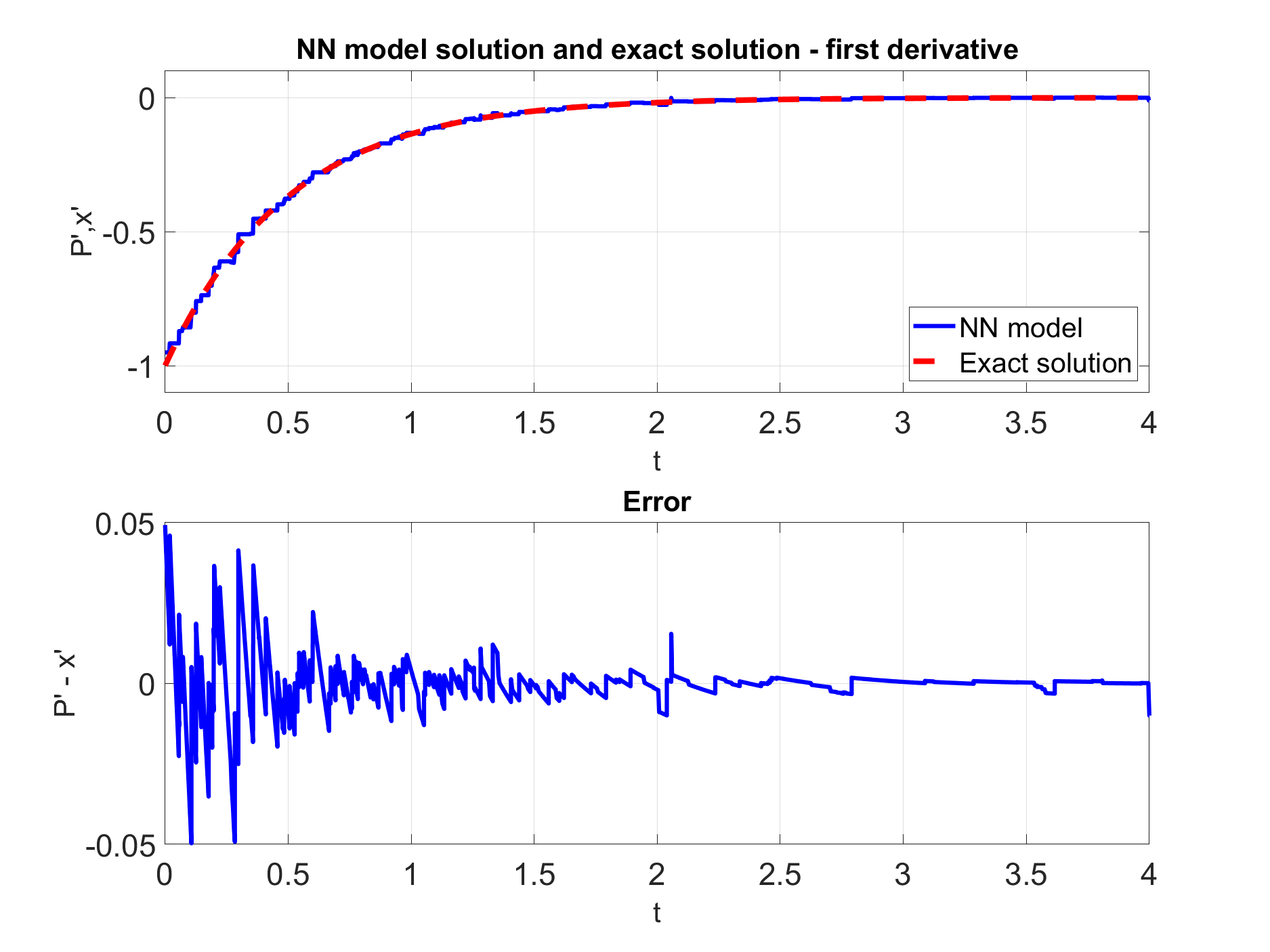}
        \caption{$x'(t)$}
    \end{subfigure}\hfill
    \begin{subfigure}[t]{0.32\linewidth}
        \centering
        \includegraphics[width=\linewidth]{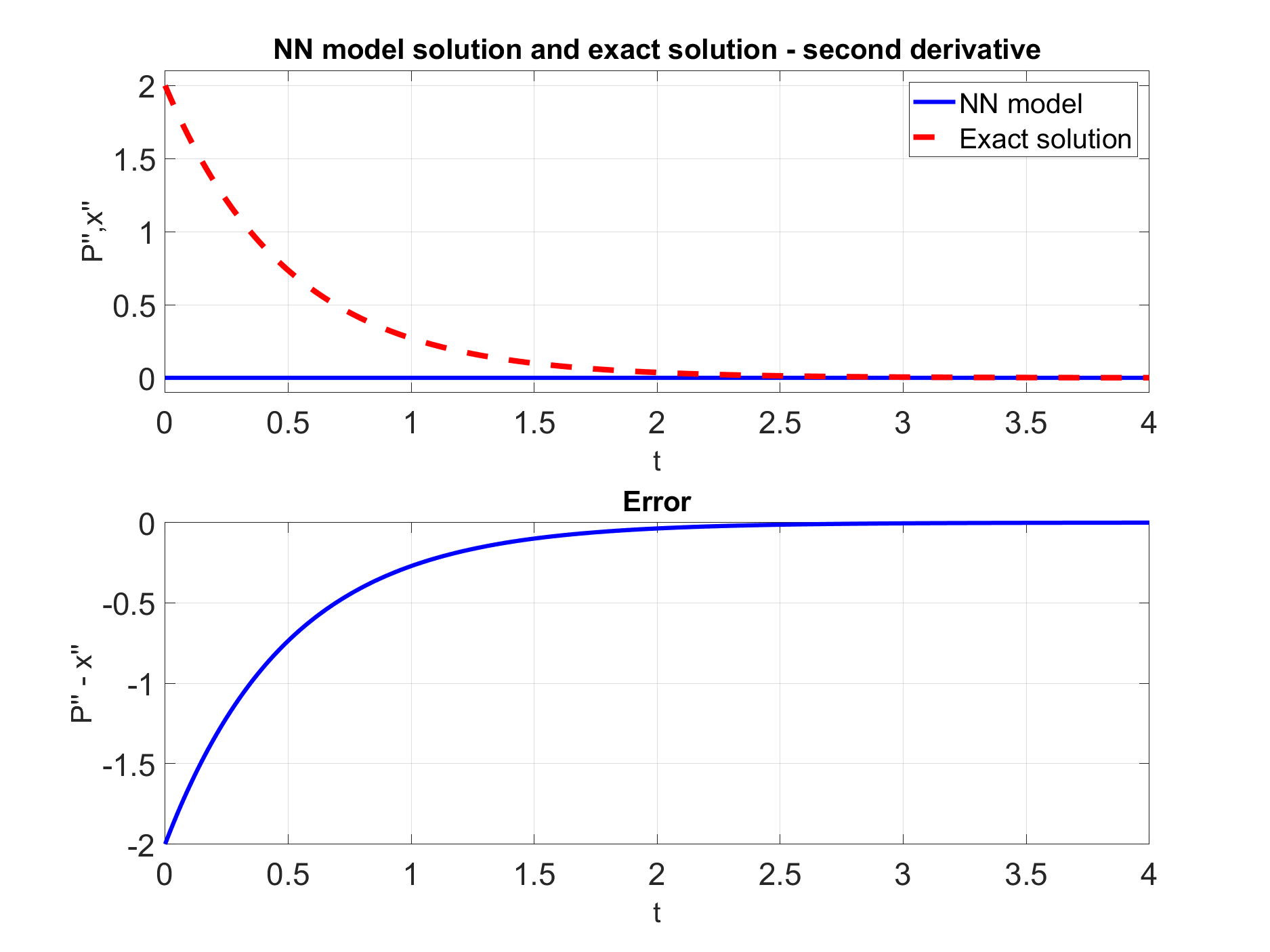}
        \caption{$x''(t)$}
    \end{subfigure}

    \vspace{10pt}

    \textbf{MLP (sigmoid)}\\[-2pt]
    \vspace*{5pt}
    \begin{subfigure}[t]{0.32\linewidth}
        \centering
        \includegraphics[width=\linewidth]{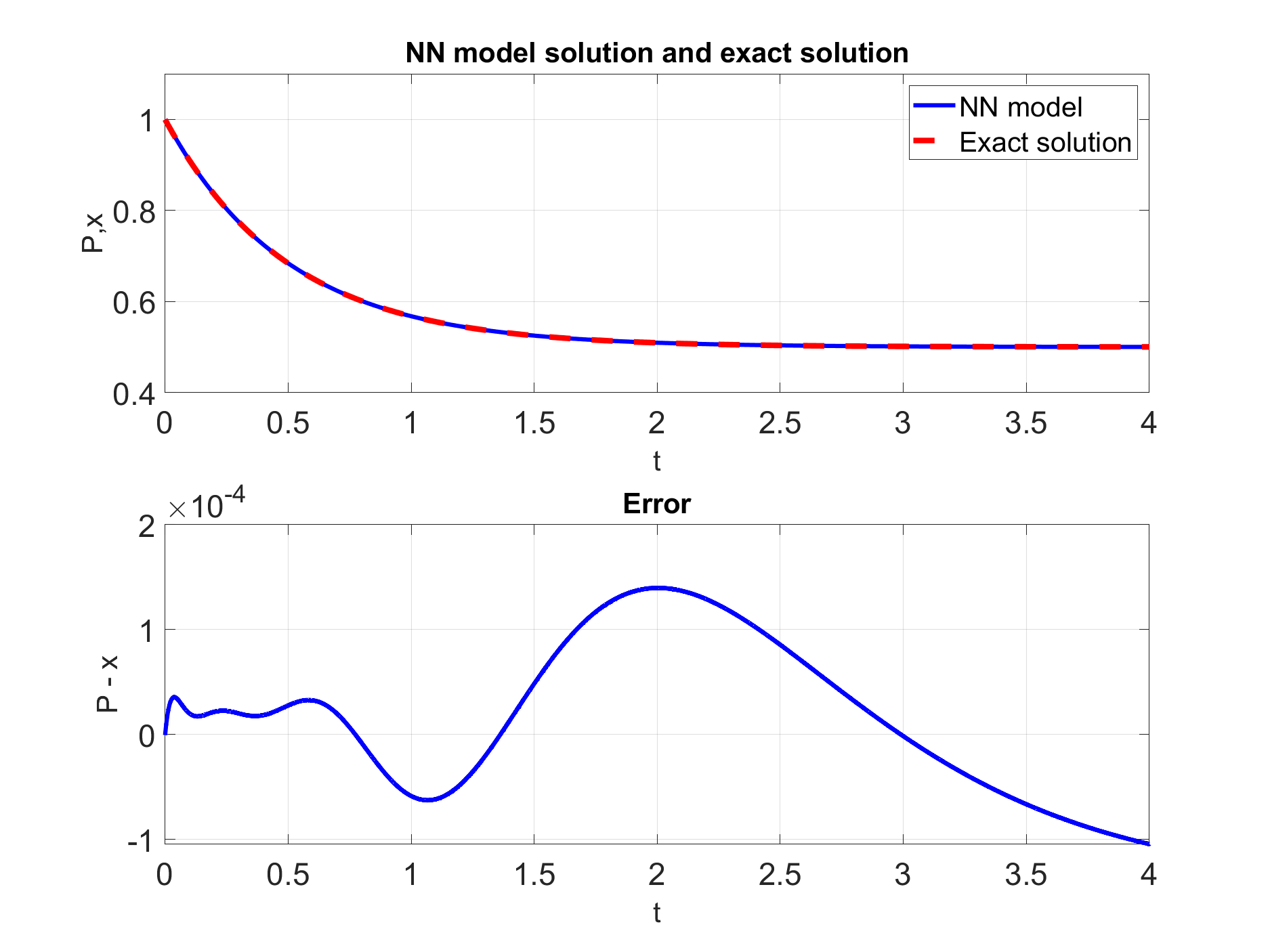}
        \caption{$x(t)$}
    \end{subfigure}\hfill
    \begin{subfigure}[t]{0.32\linewidth}
        \centering
        \includegraphics[width=\linewidth]{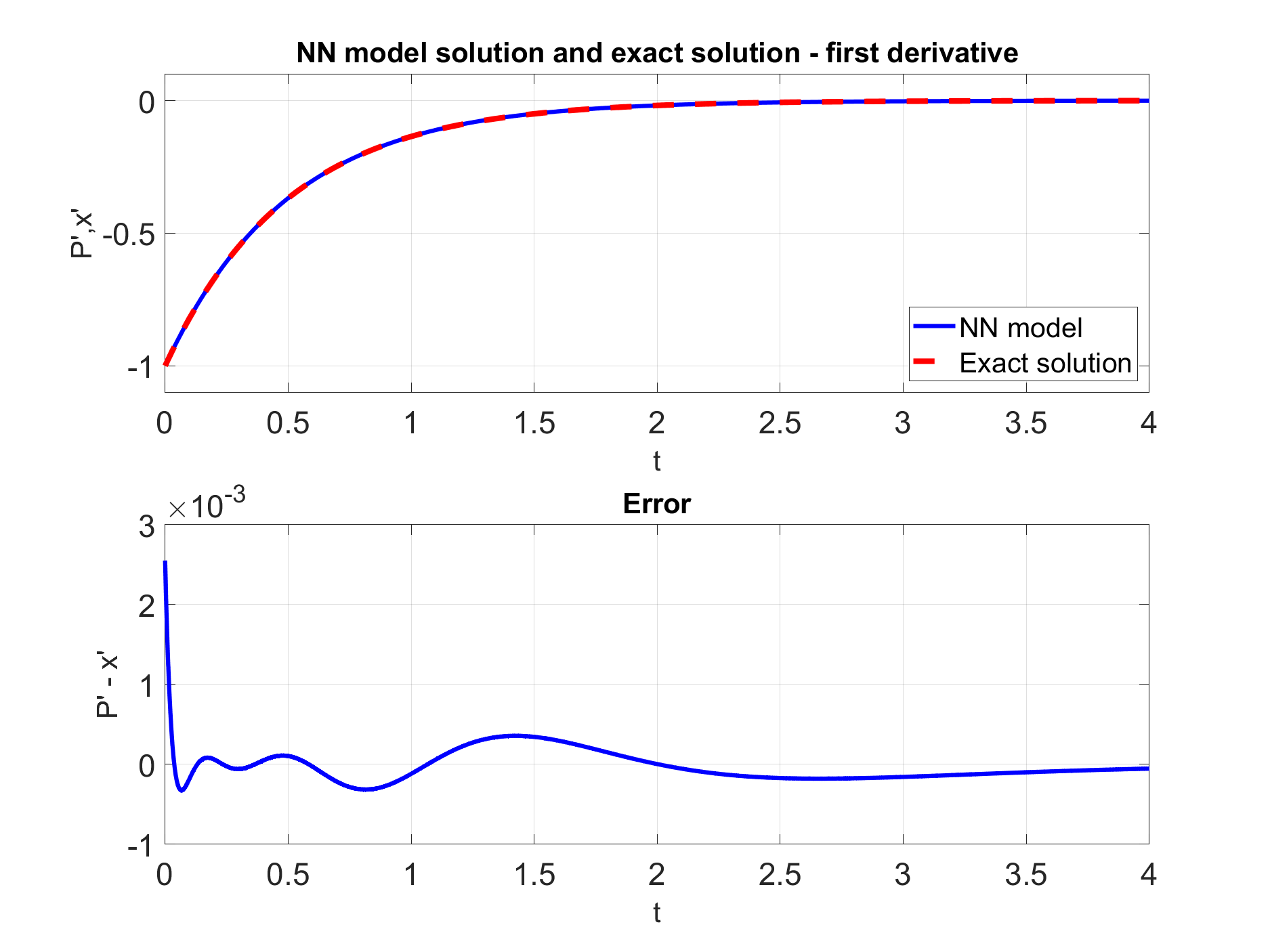}
        \caption{$x'(t)$}
    \end{subfigure}\hfill
    \begin{subfigure}[t]{0.32\linewidth}
        \centering
        \includegraphics[width=\linewidth]{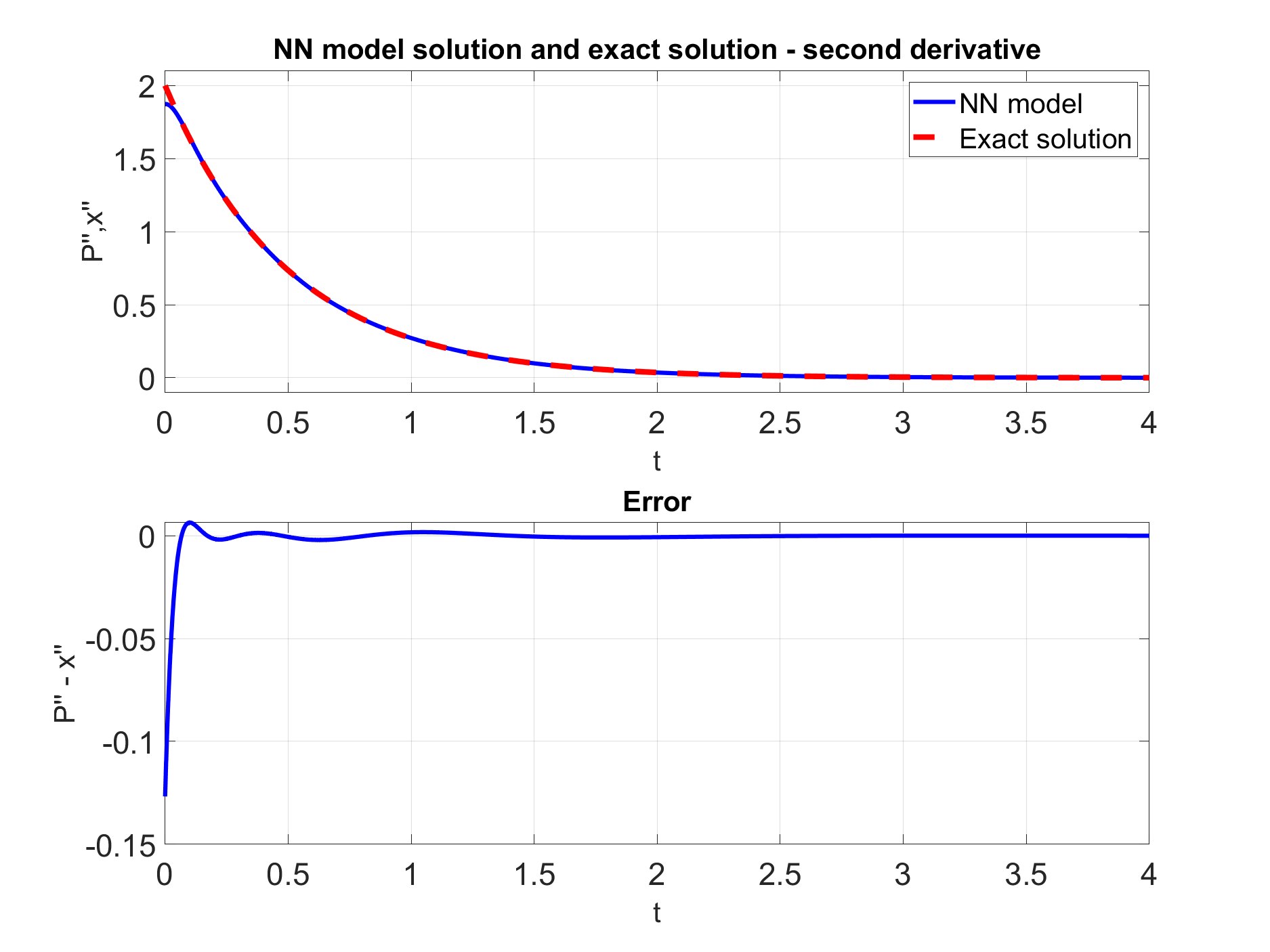}
        \caption{$x''(t)$}
    \end{subfigure}

    \vspace{10pt}

    \textbf{SIREN}\\[-2pt]
    \vspace{7pt}
    \begin{subfigure}[t]{0.32\linewidth}
        \centering
        \includegraphics[width=\linewidth]{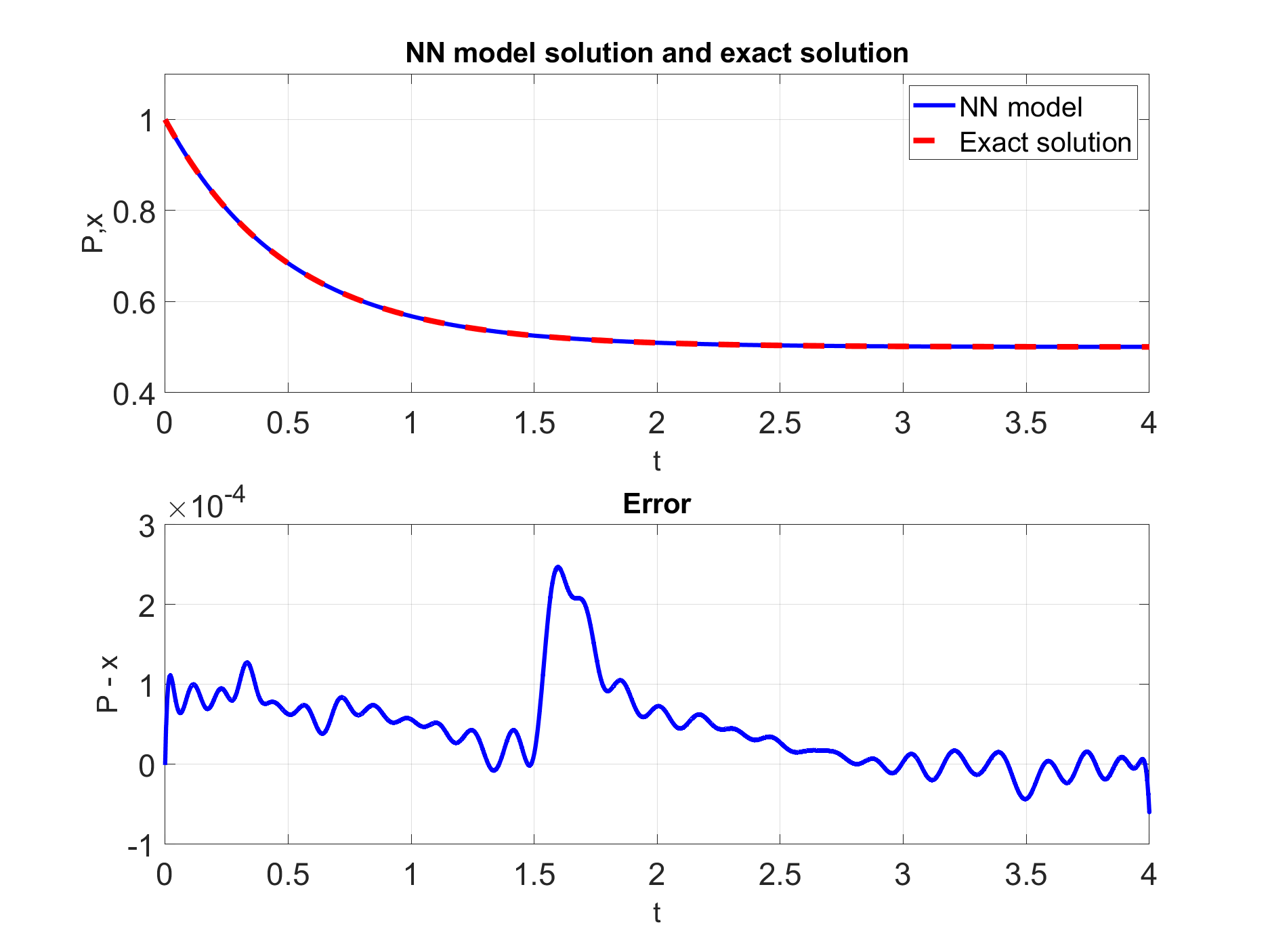}
        \caption{$x(t)$}
    \end{subfigure}\hfill
    \begin{subfigure}[t]{0.32\linewidth}
        \centering
        \includegraphics[width=\linewidth]{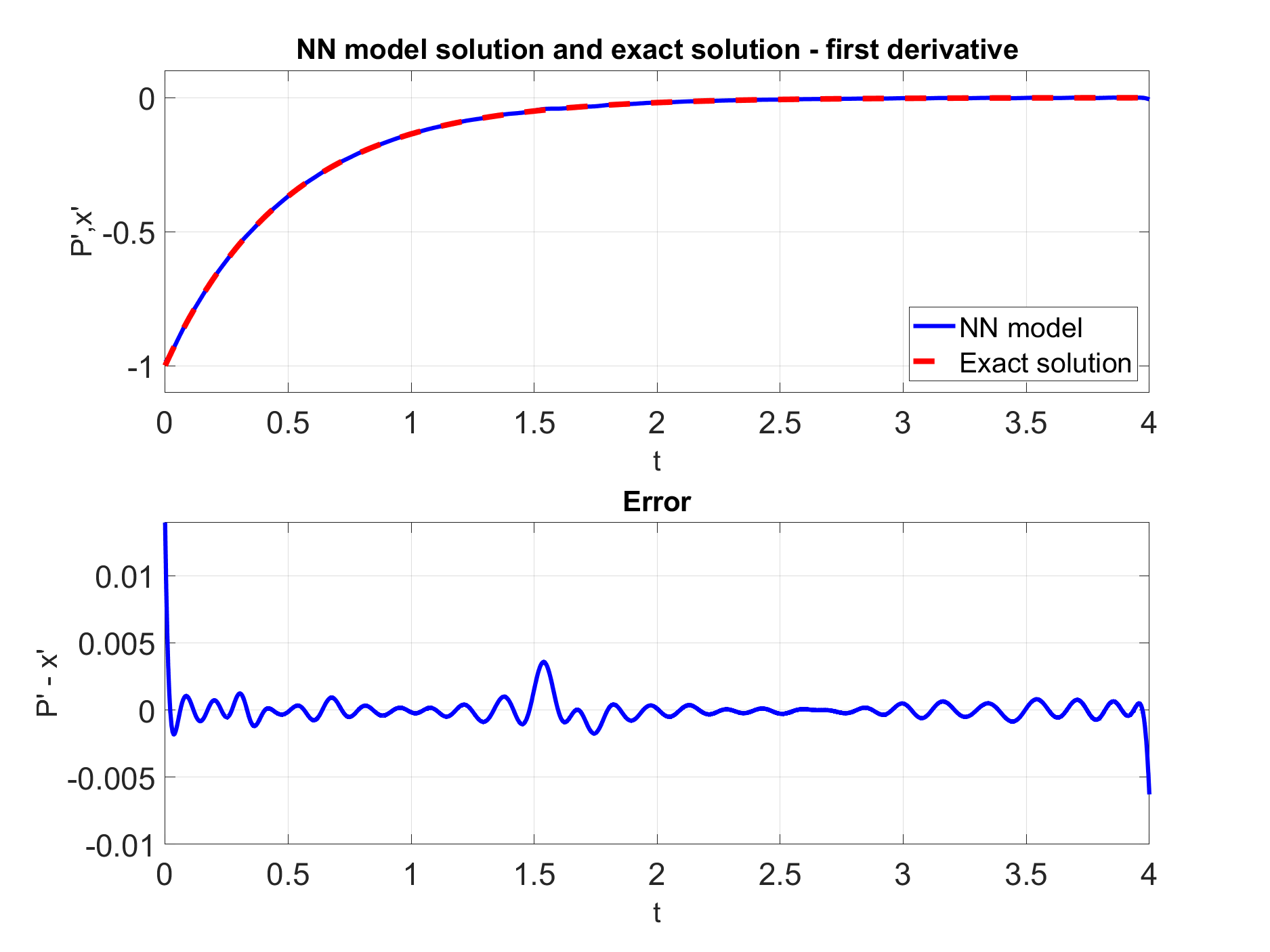}
        \caption{$x'(t)$}
    \end{subfigure}\hfill
    \begin{subfigure}[t]{0.32\linewidth}
        \centering
        \includegraphics[width=\linewidth]{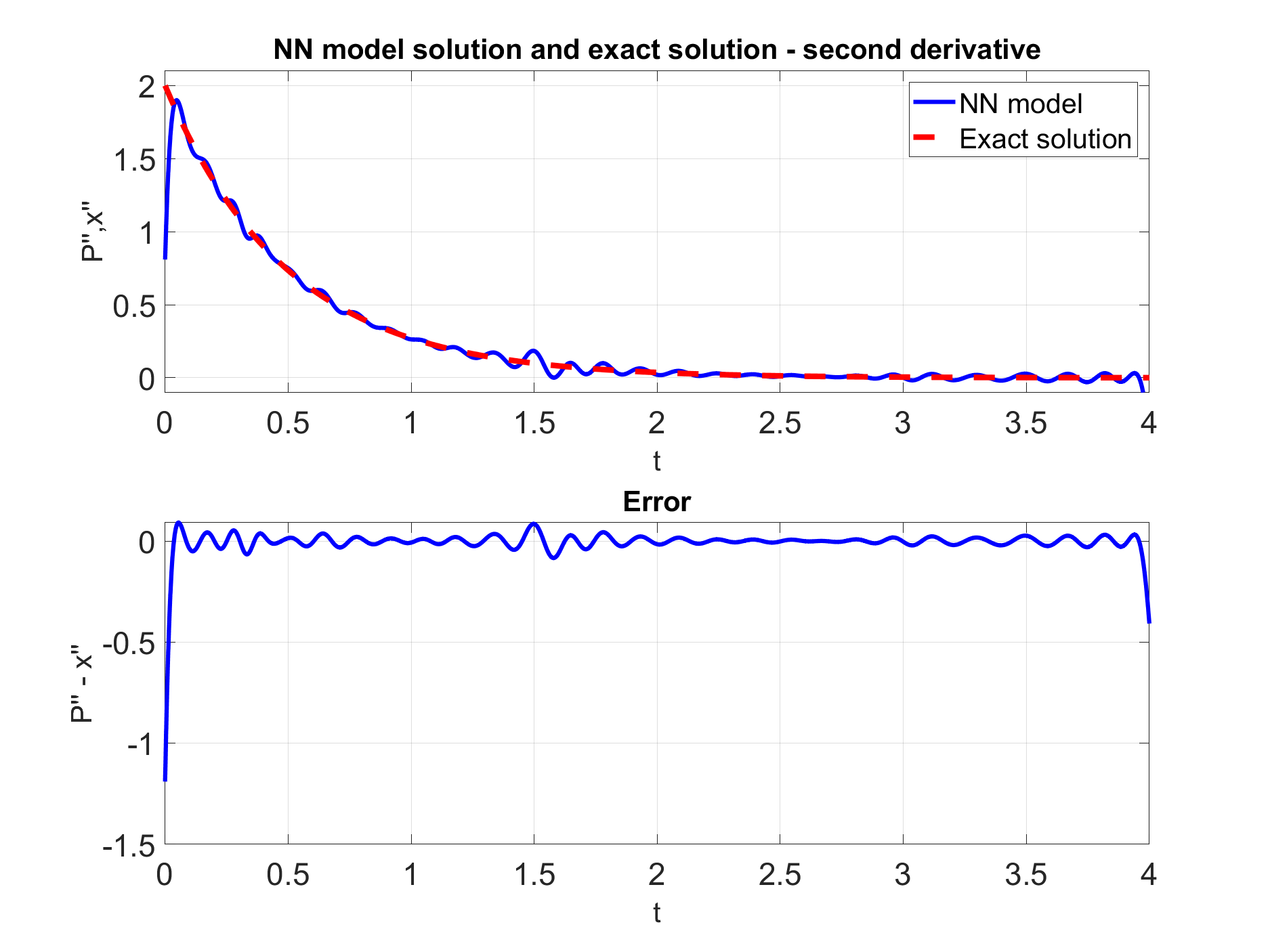}
        \caption{$x''(t)$}
    \end{subfigure}

    \caption{Type~A benchmark: baseline network predictions for the solution and its derivatives. Rows correspond to architectures (MLP with Leaky ReLU, MLP with sigmoid, and SIREN), while columns show $x(t)$, $x'(t)$, and $x''(t)$. Each panel compares the learned output against the analytical reference.}
    \label{fig:baseline_typeA}
\end{figure}

Figure~\ref{fig:baseline_typeA} compares the learned solution for Type~A and its first two derivatives against the analytical ground truth. The wide Leaky-ReLU network yields visibly larger errors, particularly in higher derivatives. This behavior is consistent with the fact that piecewise-linear activations provide limited smoothness, and their higher-order derivatives vanish almost everywhere, which is unfavorable when the training objective involves derivatives of $N(t)$ through the residual.
In contrast, the sigmoid and SIREN baselines (both with 106 parameters) provide substantially better agreement and will serve as the main reference baselines in the remainder of the paper.

The above results show that standard coordinate MLP/SIREN models can solve simple ODE instances and can achieve reasonable accuracy even with a modest number of parameters. However, their performance depends strongly on the choice of activation function and, crucially, on the tuning of penalty weights $\{\lambda_j\}$ used to enforce initial conditions via the loss. This motivates architectures that (i) restrict the hypothesis class to a structured, low-dimensional family and (ii) satisfy essential constraints \emph{exactly by construction}, thereby reducing hyperparameter sensitivity while improving accuracy per parameter.

\section{Motivation: Linear Regression Viewpoint}
\label{sec:motivation_lr}
As motivation for the neural network architecture developed in Section~V, we show how the solution of a linear ODE with constant coefficients can be approximated using a low-dimensional polynomial regression model. This viewpoint highlights two ideas that will be reused later: (i) accurate approximation on an interval of interest $I$ can be achieved with a small number of parameters, and (ii) initial conditions can be enforced \emph{exactly} by fixing a subset of coefficients (a hard-constraint construction).

\subsection{Linear ODE with constant coefficients}
We consider the $n$-th order linear ODE
\begin{equation}
    \sum_{i=0}^{n} a_i\,x^{(i)}(t) = f(t), \qquad t\in I,
    \label{eq:LODJ}
\end{equation}
where $a_i\in\mathbb{R}$ are constant coefficients and $f(t)$ is a known forcing term.

\subsection{Polynomial regression model}
We approximate $x(t)$ by a degree-$m$ polynomial model
\begin{equation}
    P(t) = \sum_{j=0}^{m} c_j\,\frac{t^j}{j!},
    \label{eq:PolyModel}
\end{equation}
where $c_j$ are coefficients to be determined. The factorial scaling is chosen so that derivatives at $t=0$ correspond directly to coefficients, which will allow exact embedding of initial conditions.

The derivatives of $P(t)$ are
\begin{equation}
    P^{(l)}(t) = \sum_{j=l}^{m} c_j\,\frac{t^{j-l}}{(j-l)!}, \qquad l=0,1,2,\ldots
    \label{eq:PolyModelDiff}
\end{equation}
Assume $m\ge n$, i.e., the polynomial degree is at least the order of the differential equation.

\subsection{Exact embedding of initial conditions}
We impose $n$ initial conditions at $t=0$:
\[
x^{(i)}(0)=x_i,\qquad i=0,1,\ldots,n-1.
\]
Evaluating \eqref{eq:PolyModelDiff} at $t=0$ yields
\[
P^{(i)}(0)=c_i,\qquad i=0,1,\ldots,n-1,
\]
and therefore the initial conditions are enforced \emph{exactly} by setting
\begin{equation}
c_i = x_i,\qquad i=0,1,\ldots,n-1.
\label{eq:ICembed_poly}
\end{equation}
Hence only the remaining coefficients $\{c_j\}_{j=n}^{m}$ are unknown.

\subsection{Substitution into the ODE and separation of known/unknown terms}
Substituting the polynomial model into \eqref{eq:LODJ} gives
\begin{align}
f(t)
&= \sum_{i=0}^{n} a_i\,P^{(i)}(t)
= \sum_{i=0}^{n} a_i \sum_{j=i}^{m} c_j \frac{t^{j-i}}{(j-i)!} \nonumber\\
&= \sum_{i=0}^{n} a_i \left(\sum_{j=i}^{n-1} c_j \frac{t^{j-i}}{(j-i)!}\right)
+ \sum_{i=0}^{n} a_i \left(\sum_{j=n}^{m} c_j \frac{t^{j-i}}{(j-i)!}\right).
\label{eq:Sol1}
\end{align}
The first term in \eqref{eq:Sol1} depends only on $c_0,\ldots,c_{n-1}$, which are fixed by \eqref{eq:ICembed_poly}. We therefore define the IC-corrected forcing
\begin{equation}
f_1(t)\coloneqq f(t)-\sum_{i=0}^{n} a_i \sum_{j=i}^{n-1} c_j \frac{t^{j-i}}{(j-i)!},
\label{eq:u1_def}
\end{equation}
so that the remaining unknown coefficients satisfy
\begin{equation}
f_1(t)=\sum_{i=0}^{n} a_i \sum_{j=n}^{m} c_j \frac{t^{j-i}}{(j-i)!}
=\sum_{j=n}^{m}\left(\sum_{i=0}^{n} a_i \frac{t^{j-i}}{(j-i)!}\right)c_j.
\label{eq:Sol2}
\end{equation}

\subsection{Collocation and least-squares estimation}
Let $\{t_k\}_{k=1}^{M}\subset I$ be collocation points and assume
\[
M > m-n+1,
\]
so the system is overdetermined. Evaluating \eqref{eq:Sol2} at $t_k$ gives
\begin{equation}
f_1(t_k)=\sum_{j=n}^{m}\left(\sum_{i=0}^{n} a_i \frac{t_k^{\,j-i}}{(j-i)!}\right)c_j,\qquad k=1,\ldots,M.
\label{eq:System1}
\end{equation}
This can be written in matrix form as
\begin{equation}
\mathbf{b}=\mathbf{A}\mathbf{c},
\label{eq:ls_system}
\end{equation}
where $\mathbf{c}=[c_n,\ldots,c_m]^\top\in\mathbb{R}^{m-n+1}$, $\mathbf{b}\in\mathbb{R}^{M}$, $\mathbf{A}\in\mathbb{R}^{M\times(m-n+1)}$, and
\[
b_k = f_1(t_k),\qquad
A_{k,\,(j-n+1)}=\sum_{i=0}^{n} a_i \frac{t_k^{\,j-i}}{(j-i)!},\qquad j=n,\ldots,m.
\]

We estimate the unknown coefficients by least squares:
\begin{equation}
\hat{\mathbf{c}}=\arg\min_{\mathbf{c}}\|\mathbf{A}\mathbf{c}-\mathbf{b}\|_2^2 .
\label{eq:LS}
\end{equation}
When $\mathbf{A}$ has full column rank, the minimizer can be expressed via the normal equations
\begin{equation}
\hat{\mathbf{c}} = (\mathbf{A}^\top \mathbf{A})^{-1}\mathbf{A}^\top \mathbf{b},
\label{eq:normal_eq}
\end{equation}
although in practice $\hat{\mathbf{c}}$ is preferably computed using a QR- or SVD-based least-squares solver for improved numerical stability.

Finally, combining $\hat{\mathbf{c}}$ with the fixed coefficients $c_0,\ldots,c_{n-1}$ from \eqref{eq:ICembed_poly} yields the polynomial approximation
\[
P(t)=\sum_{j=0}^{m} c_j \frac{t^j}{j!},
\]
which serves as a low-parameter model of the solution of \eqref{eq:LODJ} on $I$.

\subsection{Implications for parameter-minimal neural solvers}
This regression construction demonstrates that (i) the solution can be represented using a small number of degrees of freedom and (ii) essential constraints (initial conditions) can be enforced exactly by fixing low-order coefficients.
However, global polynomials may require higher degree over longer intervals and do not directly extend to more flexible function classes.
Section~V retains the same two principles---low-dimensional parameterization and hard initial-condition embedding---but implements them as a trainable neural module using a nested (Horner) form and introduces extensions that improve approximation power while keeping the number of learnable parameters small.

\subsection{Example 1 (Type~A)}
We first consider the Type~A benchmark ODE from Table~I and solve it using the polynomial regression model described above. We use a degree-$m=15$ polynomial (i.e., $m+1=16$ coefficients) and $M=10{,}000$ collocation samples $\{t_k\}_{k=1}^{M}$ drawn i.i.d.\ from the uniform distribution on the interval of interest $I=[0,4]$. The corresponding values $f(t_k)$ are computed from the known forcing term, and the IC-corrected samples $f_1(t_k)$ are formed according to \eqref{eq:u1_def}.

\begin{figure}[t]
    \centering
    \begin{subfigure}[t]{0.48\linewidth}
        \centering
        \includegraphics[width=\linewidth]{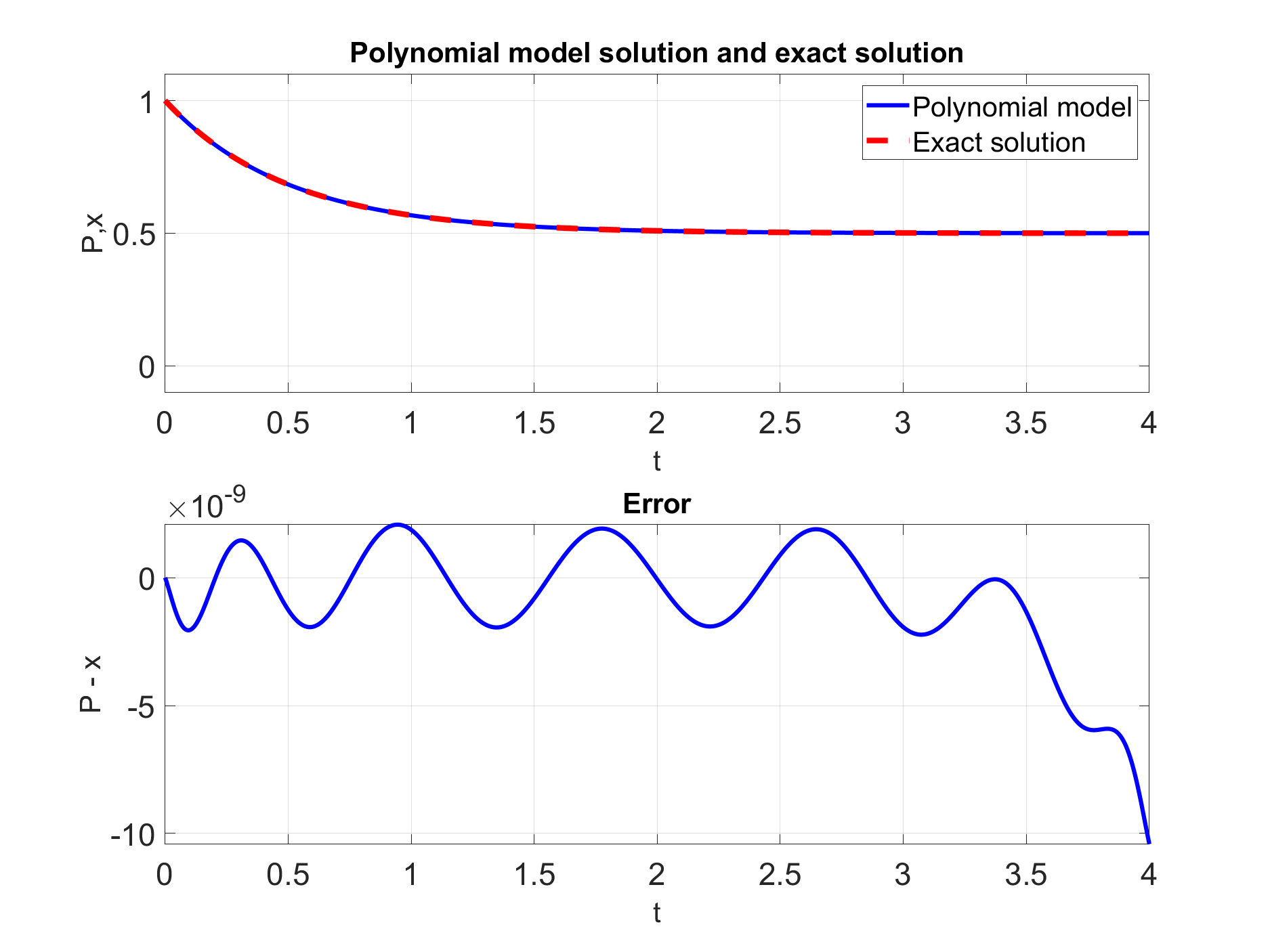}
        \caption{Solution $x(t)$ and error.}
        \label{fig:Ex1_f}
    \end{subfigure}\hfill
    \begin{subfigure}[t]{0.48\linewidth}
        \centering
        \includegraphics[width=\linewidth]{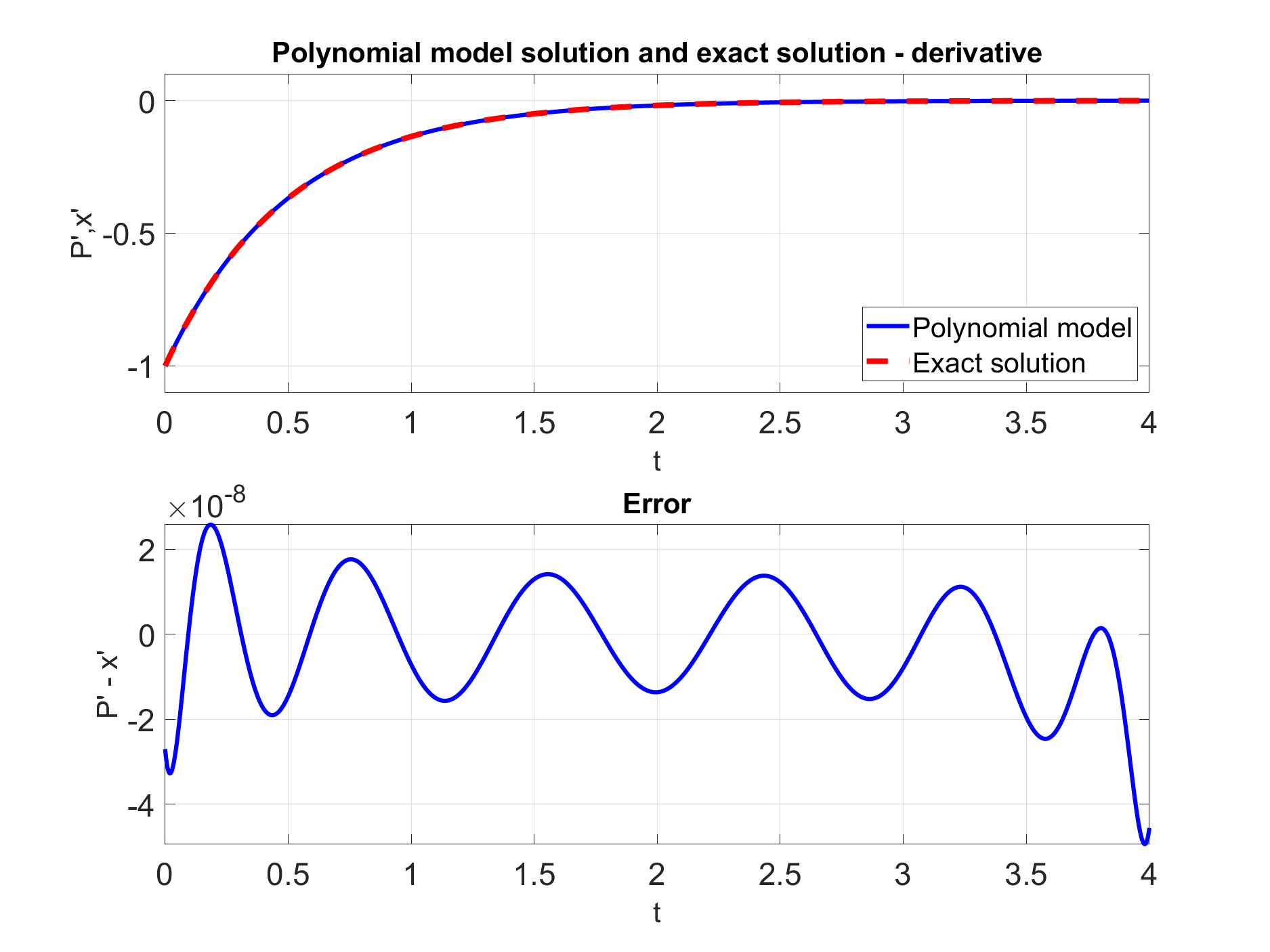}
        \caption{Derivative $x'(t)$ and error.}
        \label{fig:Ex1_fd}
    \end{subfigure}
    \caption{Type~A: polynomial regression (degree $m=15$) versus the analytical solution on $I=[0,4]$.}
    \label{fig:Ex1}
\end{figure}

Figure~\ref{fig:Ex1} compares the polynomial approximation with the analytical solution, and also reports the corresponding errors. Despite using only 16 parameters, the regression model provides a very accurate approximation of the solution and its derivatives on $I$.

\subsection{Example 2 (matched forcing)}
Next, we consider the ODE
\[
\begin{aligned}
x'(t) + 2x(t) &= e^{-2t},\\
x(0) &= 0,
\end{aligned}
\]
whose closed-form solution is $x(t)=t e^{-2t}$. This example is intentionally chosen so that the forcing term matches the decay rate of the homogeneous solution, which yields a polynomially modulated exponential response.

\begin{figure}[t]
    \centering
    \begin{subfigure}[t]{0.48\linewidth}
        \centering
        \includegraphics[width=\linewidth]{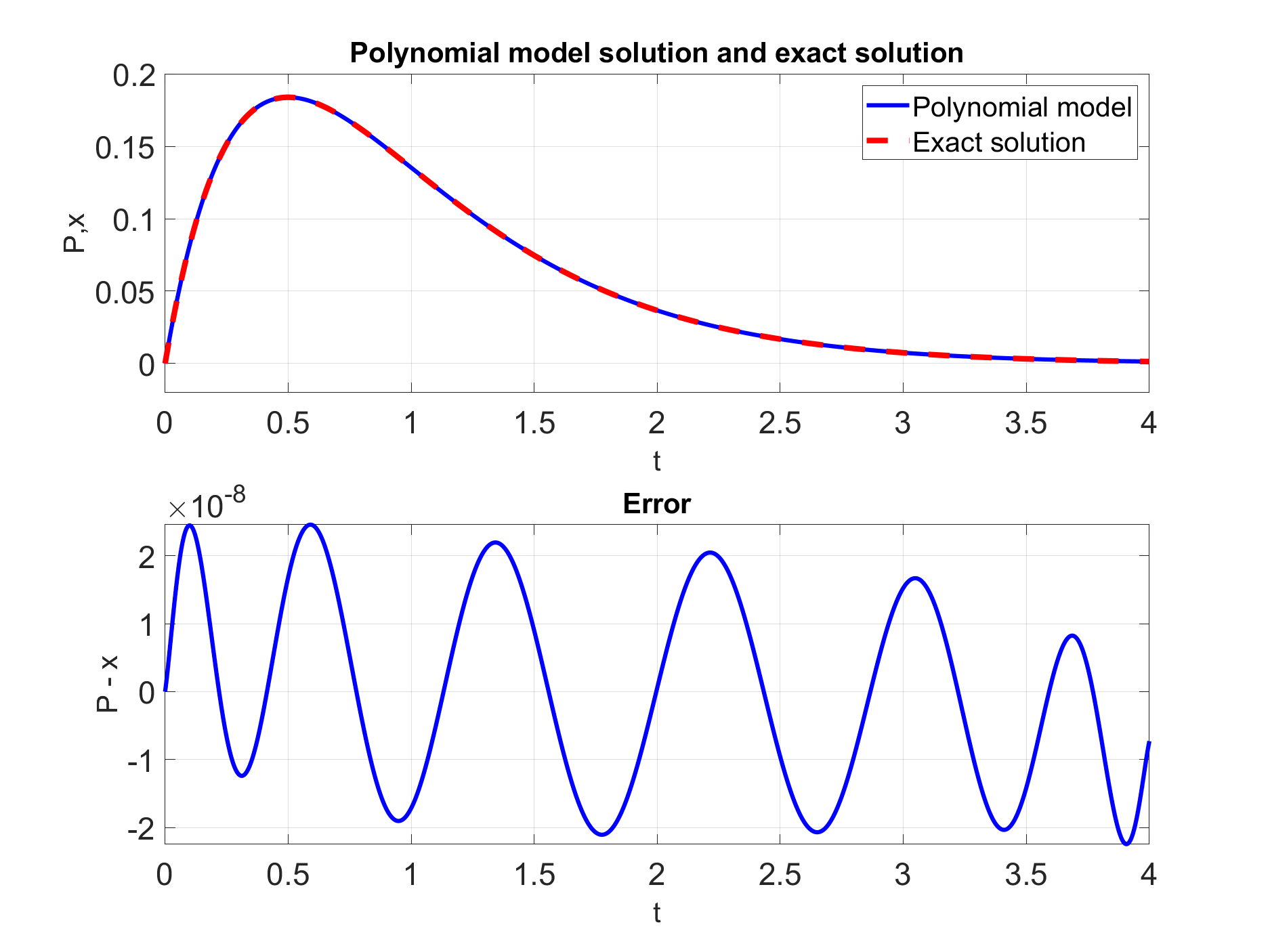}
        \caption{Solution $x(t)$ and error.}
        \label{fig:Ex2_f}
    \end{subfigure}\hfill
    \begin{subfigure}[t]{0.48\linewidth}
        \centering
        \includegraphics[width=\linewidth]{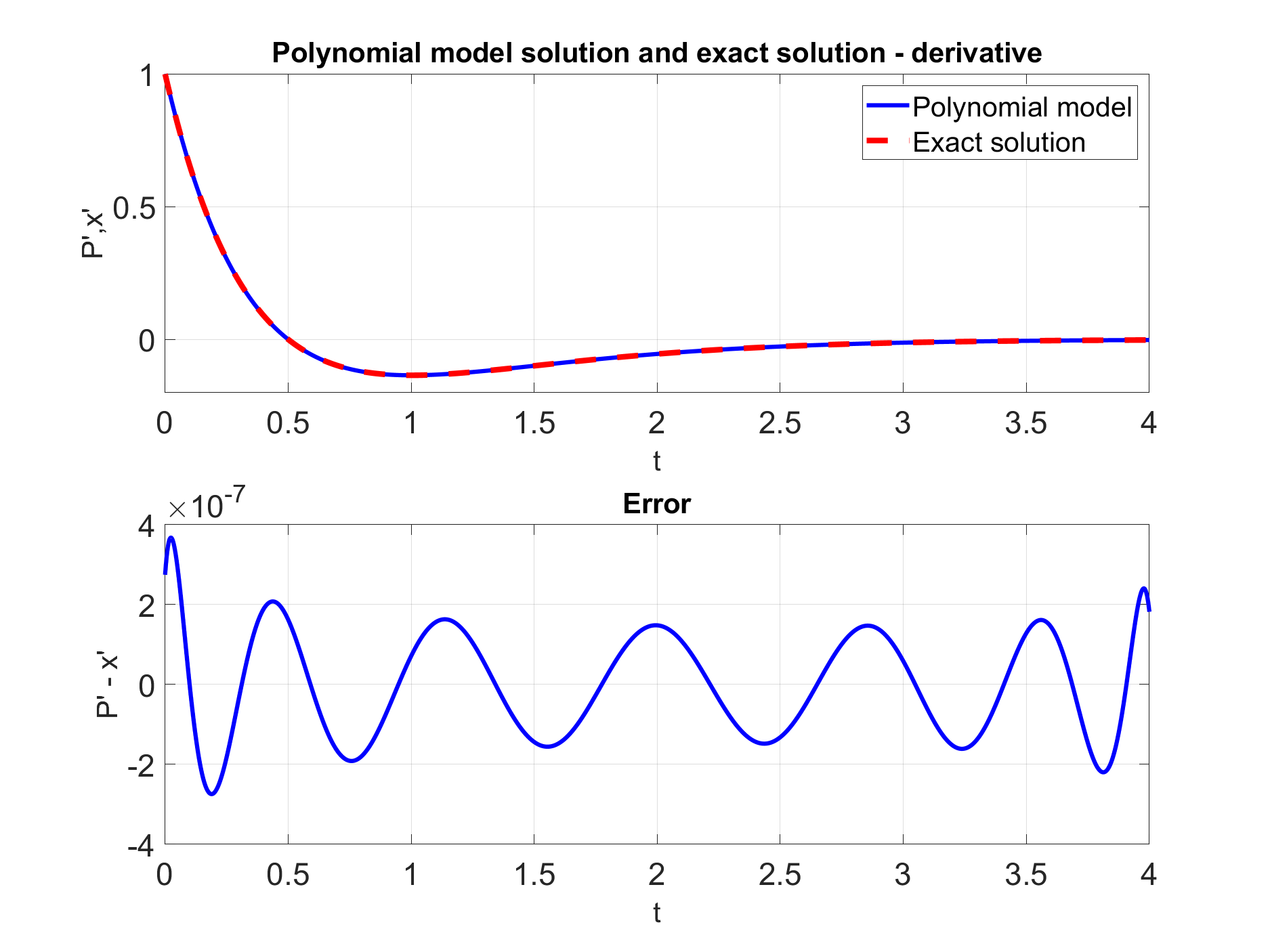}
        \caption{Derivative $x'(t)$ and error.}
        \label{fig:Ex2_fd}
    \end{subfigure}
    \caption{Matched forcing: polynomial regression (degree $m=15$) versus the analytical solution on $I=[0,4]$.}
    \label{fig:Ex2}
\end{figure}

We keep the same regression settings as in Example~1 ($m=15$, $M=10{,}000$, $t_k\sim\mathcal{U}(I)$ with $I=[0,4]$). Figure~\ref{fig:Ex2} compares the polynomial approximation to the exact solution and also shows the corresponding derivative-level errors. As in Example~1, a 16-parameter polynomial provides a strong approximation over the interval of interest.

\subsection{Example 3 (Type~C)}
\begin{figure}[t]
    \centering
    \begin{subfigure}[t]{0.32\linewidth}
        \centering
        \includegraphics[width=\linewidth]{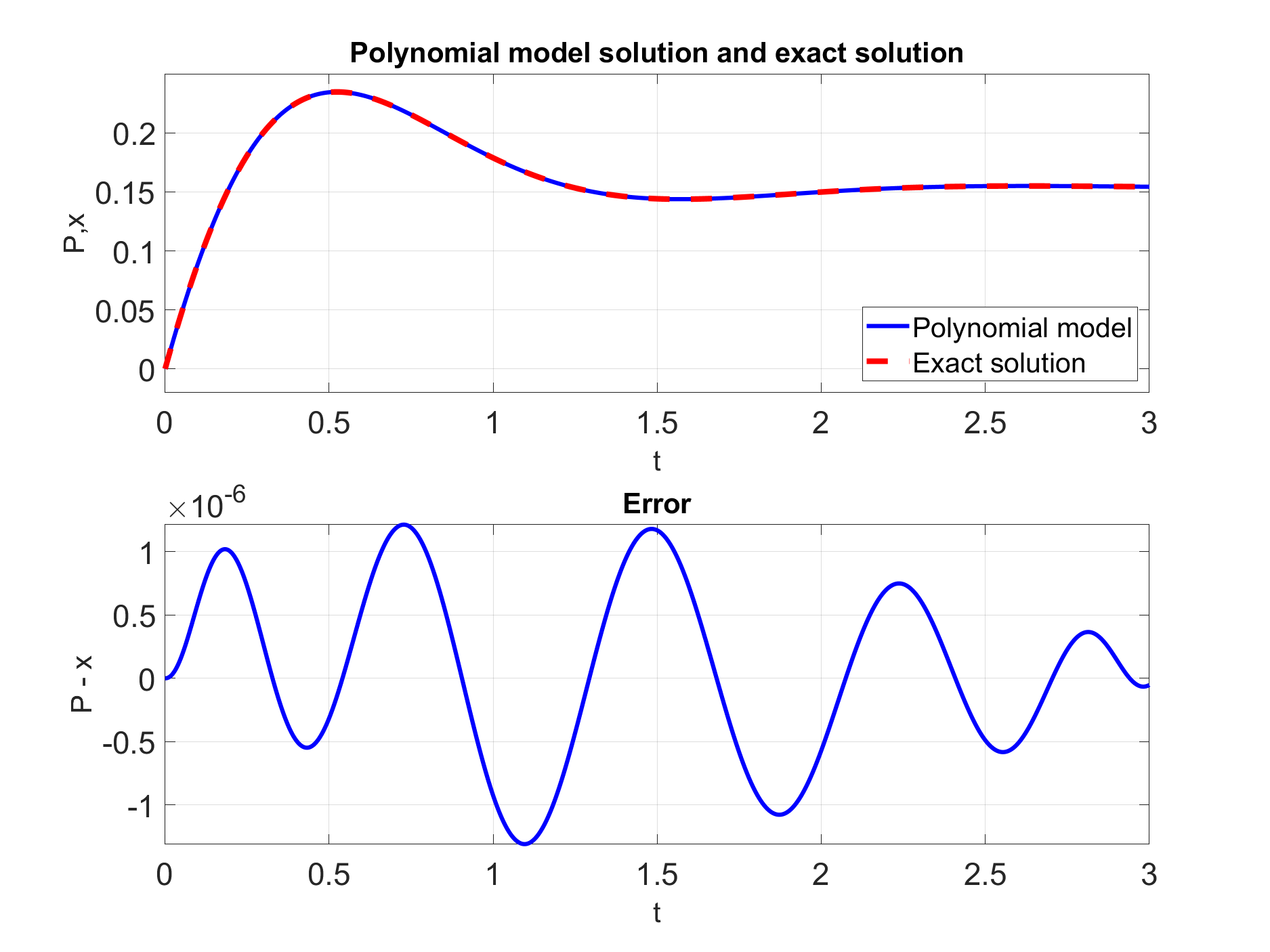}
        \caption{Solution $x(t)$ and error.}
        \label{fig:Ex3_f}
    \end{subfigure}\hfill
    \begin{subfigure}[t]{0.32\linewidth}
        \centering
        \includegraphics[width=\linewidth]{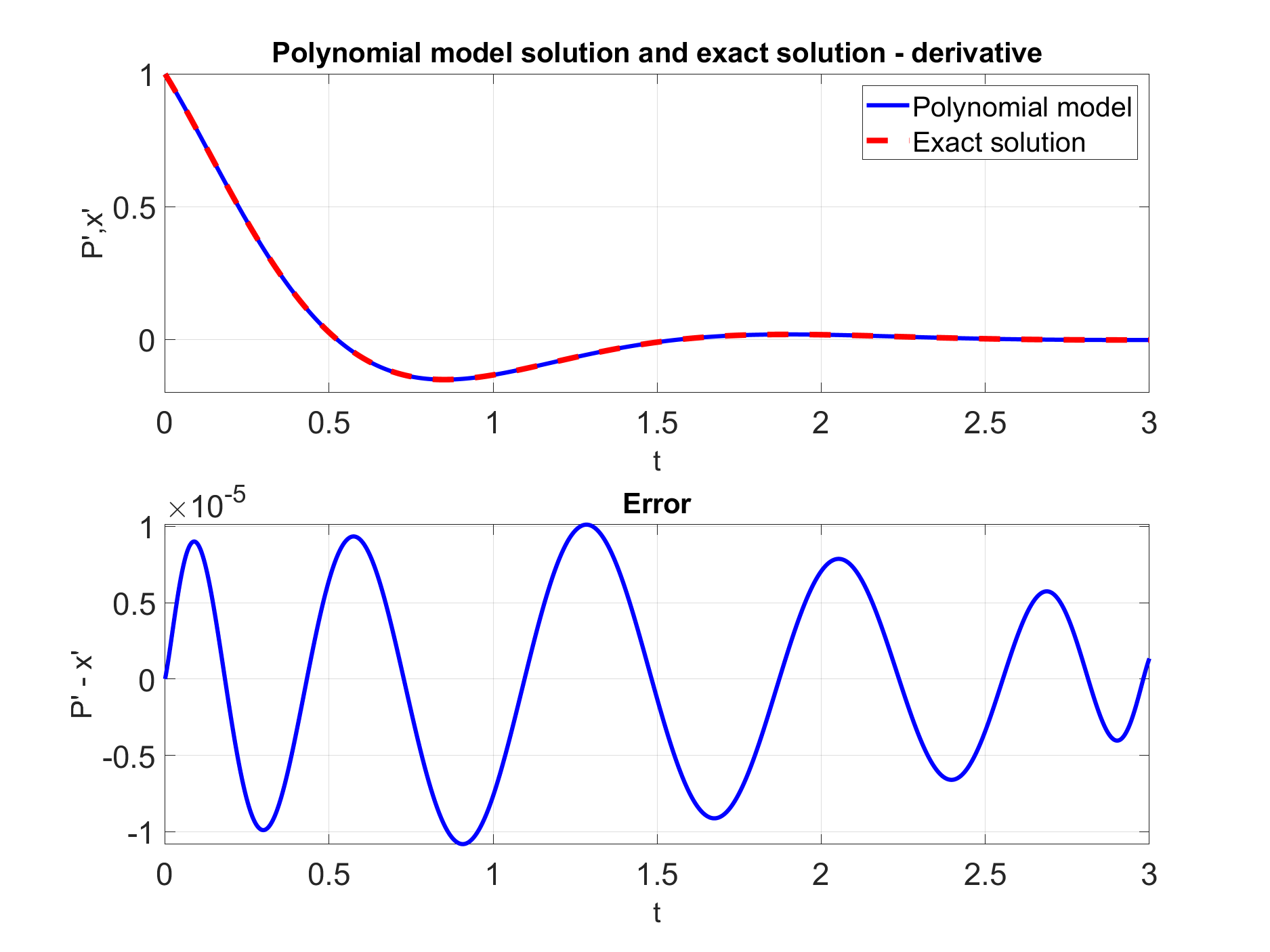}
        \caption{First derivative and error.}
        \label{fig:Ex3_fd}
    \end{subfigure}\hfill
    \begin{subfigure}[t]{0.32\linewidth}
        \centering
        \includegraphics[width=\linewidth]{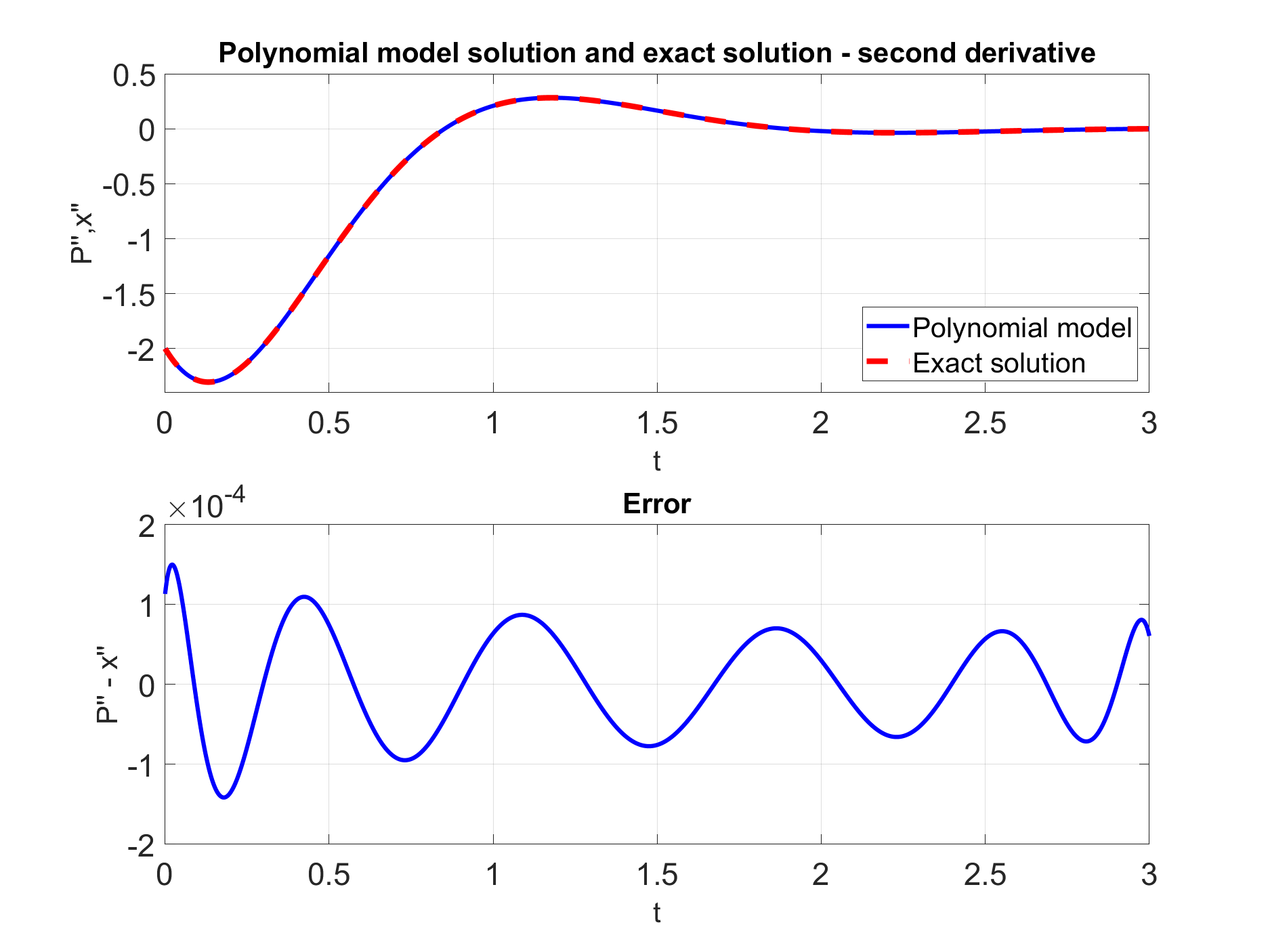}
        \caption{Second derivative and error.}
        \label{fig:Ex3_fdd}
    \end{subfigure}
    \caption{Type~C: polynomial regression (degree $m=15$) versus the analytical solution on $I=[0,3]$.}
    \label{fig:Ex3}
\end{figure}

Finally, we consider the Type~C benchmark ODE from Table~I, which exhibits damped oscillatory behavior. We again use the same regression settings: $m=15$, $M=10{,}000$, $I=[0,3]$. Figure~\ref{fig:Ex3} compares the polynomial regression model with the analytical solution and reports the errors for the solution as well as for the first and second derivatives. The approximation remains accurate across both the transient and oscillatory components.

These examples indicate that a very small number of degrees of freedom can already yield accurate DE solutions on moderate intervals when the initial conditions are embedded by construction. However, global polynomial models can become less reliable as the interval grows or the target behavior becomes more complex; in particular, higher-degree global polynomials may suffer from boundary oscillations and sensitivity to sampling (a classical issue related to Runge-type behavior). This motivates the next section: we retain the same two principles: i) a low-dimensional parameterization, and ii) exact enforcement of essential constraints but implement them via a structured, trainable neural module (Horner form) that is more flexible and can be extended (e.g., piecewise constructions) while keeping the parameter count small.

\section{Horner Networks (Parameter-Minimal Neural DE Solvers)}
\label{sec:horner_nets}
In this section, we introduce a parameter-minimal neural architecture for solving differential equations by
restricting the hypothesis class to a Horner-factorized polynomial. The resulting model is a smooth, fully
differentiable implicit representation whose learnable parameters correspond directly to polynomial degrees of
freedom. We evaluate this architecture on the benchmark ODEs from Table~I and compare it against the MLP and
SIREN baselines from Section~III.

\subsection{Horner scheme and network architecture}
\label{sec:horner_scheme}
A degree-$m$ polynomial can be evaluated efficiently in nested (Horner) form,
\begin{equation}
P_h(t)=a_0+t\Big(a_1+t\big(a_2+\cdots+t(a_{m-1}+a_m t)\cdots\big)\Big),
\label{eq:Horner}
\end{equation}
which requires only $m$ multiply--add stages and yields an efficient (and numerically stable) evaluation procedure.
Equivalently, \eqref{eq:Horner} can be written as the recursion
\[
z_m=a_m,\qquad z_i=a_i+t\,z_{i+1}\;\; (i=m-1,\ldots,0),\qquad P_h(t)=z_0.
\]
This recursion directly motivates our architecture: a \emph{Basic Horner Block} (BHB) implements one stage
$z_i=a_i+t\,z_{i+1}$ with a single coefficient $a_i$, and chaining $m{+}1$ blocks yields the full \emph{Horner Network}
(HN).

\begin{figure}[t]
    \centering
    \begin{subfigure}[t]{0.48\linewidth}
        \centering
        \includegraphics[width=\linewidth]{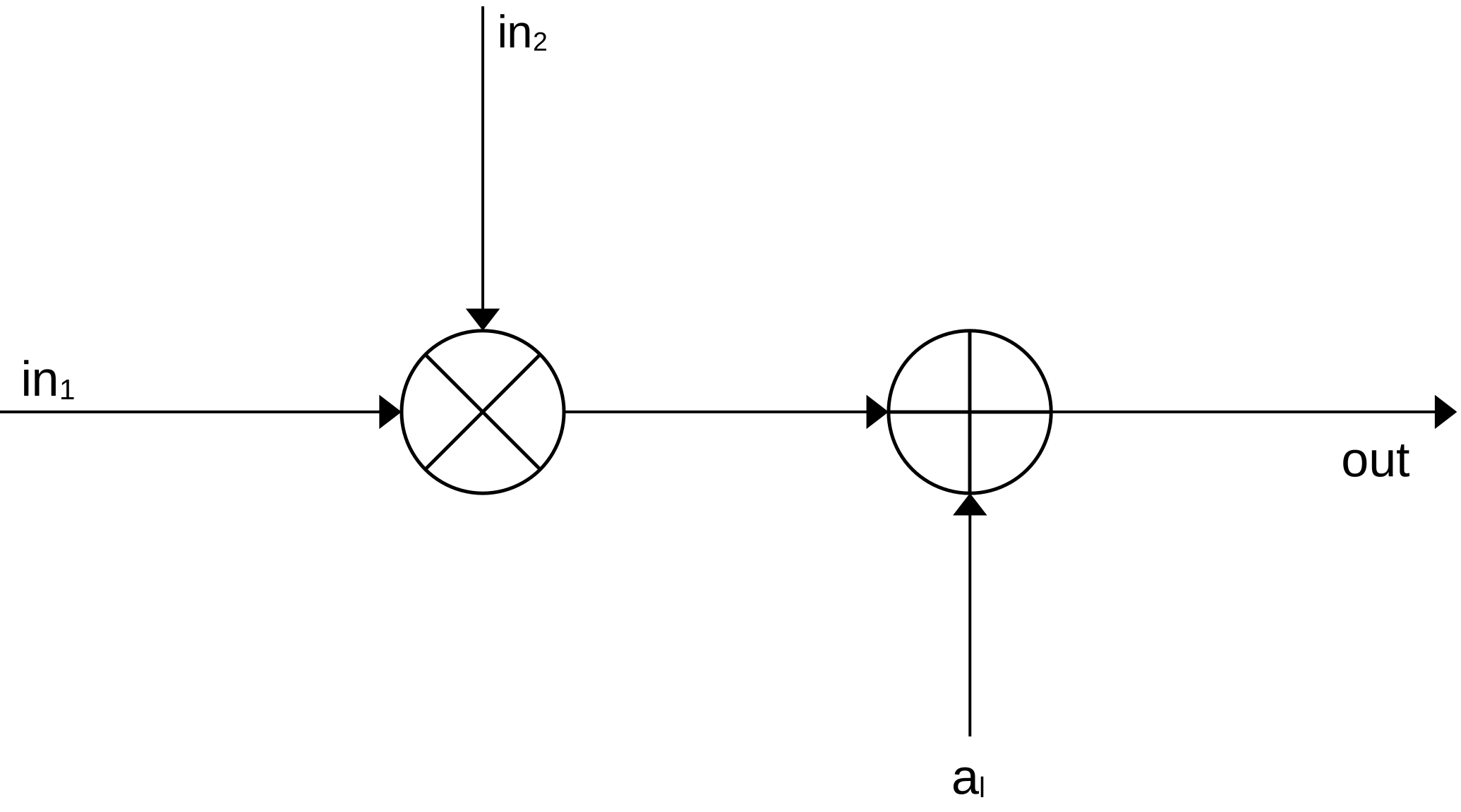}
        \caption{Basic Horner Block (BHB): one multiply--add stage with coefficient $a_i$.}
        \label{fig:BHB}
    \end{subfigure}\hfill
    \begin{subfigure}[t]{0.48\linewidth}
        \centering
        \includegraphics[width=\linewidth]{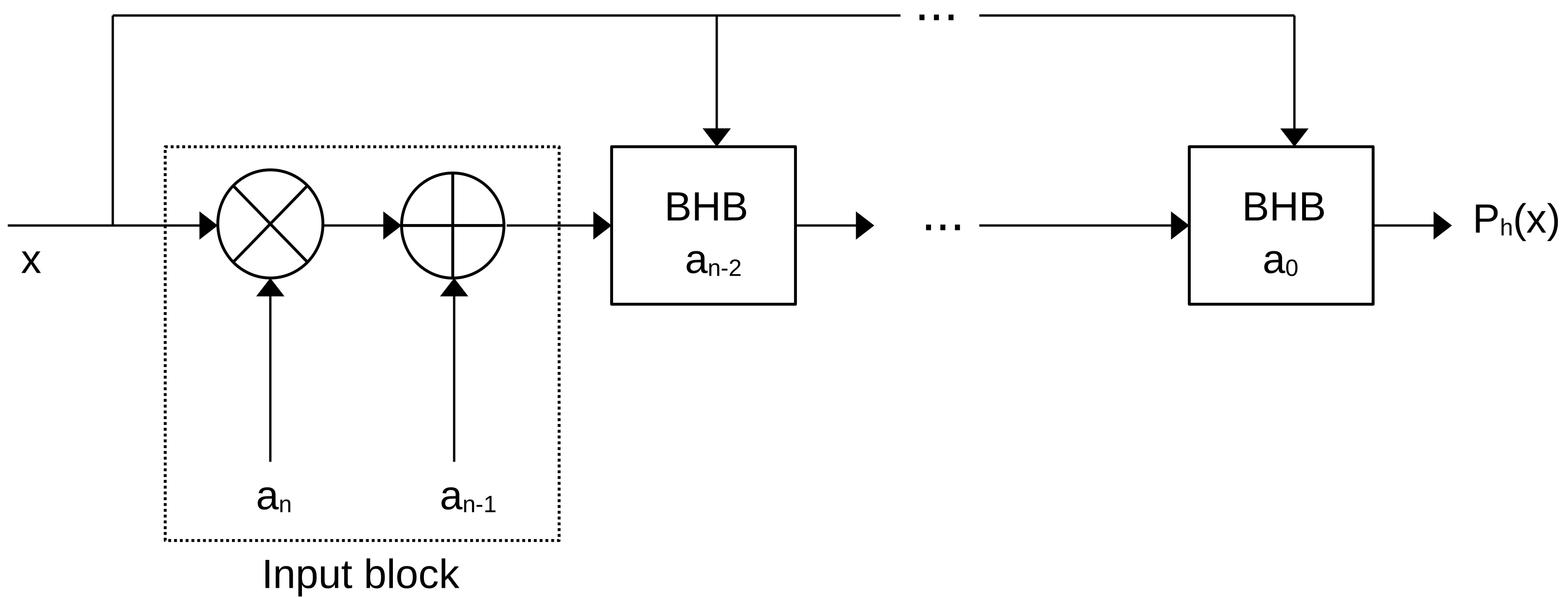}
        \caption{Horner Network (HN): cascade of BHBs implementing \eqref{eq:Horner}.}
        \label{fig:HN}
    \end{subfigure}
    \caption{Horner-based architecture used as a parameter-minimal neural solver.}
    \label{fig:BHBiHN}
\end{figure}

\subsection{Hard embedding of initial conditions}
\label{sec:horner_ic}
Let $\mathcal{N}(t)$ denote the Horner network output (i.e., $\mathcal{N}(t)=P_h(t)$). For an $n$-th order ODE with
initial conditions $x^{(j)}(0)=x_j$, $j=0,\ldots,n-1$, we embed these constraints \emph{by construction} by fixing the
corresponding low-order coefficients.
For the benchmark problems considered here, this reduces to:
\[
a_0=\mathcal{N}(0)=x(0)=x_0 \quad\text{(Types A and B)},
\]
and for the second-order case,
\[
a_0=\mathcal{N}(0)=x(0)=x_0,\qquad
a_1=\frac{d}{dt}\mathcal{N}(0)=x'(0)=x_1 \quad\text{(Type C)}.
\]
As a result, the initial conditions are satisfied exactly and do not require additional penalty terms in the loss.

\subsection{Residual loss and training protocol}
\label{sec:horner_training}
We follow the standard residual-minimization paradigm: given collocation points $T=\{t_k\}_{k=1}^{M}\subset I$ with
known forcing samples $f_k=f(t_k)$, we minimize the mean-squared residual
\begin{equation}
\mathcal{L}_{\mathrm{HN}}
=
\frac{1}{M}\sum_{k=1}^{M}
\Big(
F\!\Big(t_k,\mathcal{N}(t_k),\tfrac{d}{dt}\mathcal{N}(t_k),\ldots,\tfrac{d^n}{dt^n}\mathcal{N}(t_k)\Big)
-
f(t_k)
\Big)^2,
\label{eq:HNloss}
\end{equation}
where derivatives are obtained by automatic differentiation.

Unless stated otherwise, we use $M=200$ collocation points, Adam optimization with initial learning rate $10^{-3}$,
and train for $10{,}000$ epochs. We test the Horner network on the three benchmark ODEs (Types A--C).

To assess the accuracy of the learned model, we report the root mean square error (RMSE) between the network-predicted solution $\mathcal{N}(t)$ and the corresponding analytical solution $x(t)$ (Table~\ref{tab:bench_odes}):
\begin{equation}
\mathrm{RMSE}(\mathcal{N},x)
=
\sqrt{\frac{1}{N_{\mathrm{eval}}}\sum_{k=1}^{N_{\mathrm{eval}}}
\bigl(\mathcal{N}(\tilde t_k)-x(\tilde t_k)\bigr)^2 }.
\label{eq:rmse_solution}
\end{equation}
The error is computed on a dense uniform evaluation grid
$T_{\mathrm{eval}}=\{\tilde t_k\}_{k=1}^{N_{\mathrm{eval}}}\subseteq I$ with $N_{\mathrm{eval}}=100\,000$ points .

\subsection{Results on benchmark ODEs}
\label{sec:horner_results}
\begin{figure}[t]
    \centering
    \begin{subfigure}[t]{0.32\linewidth}
        \centering
        \includegraphics[width=\linewidth]{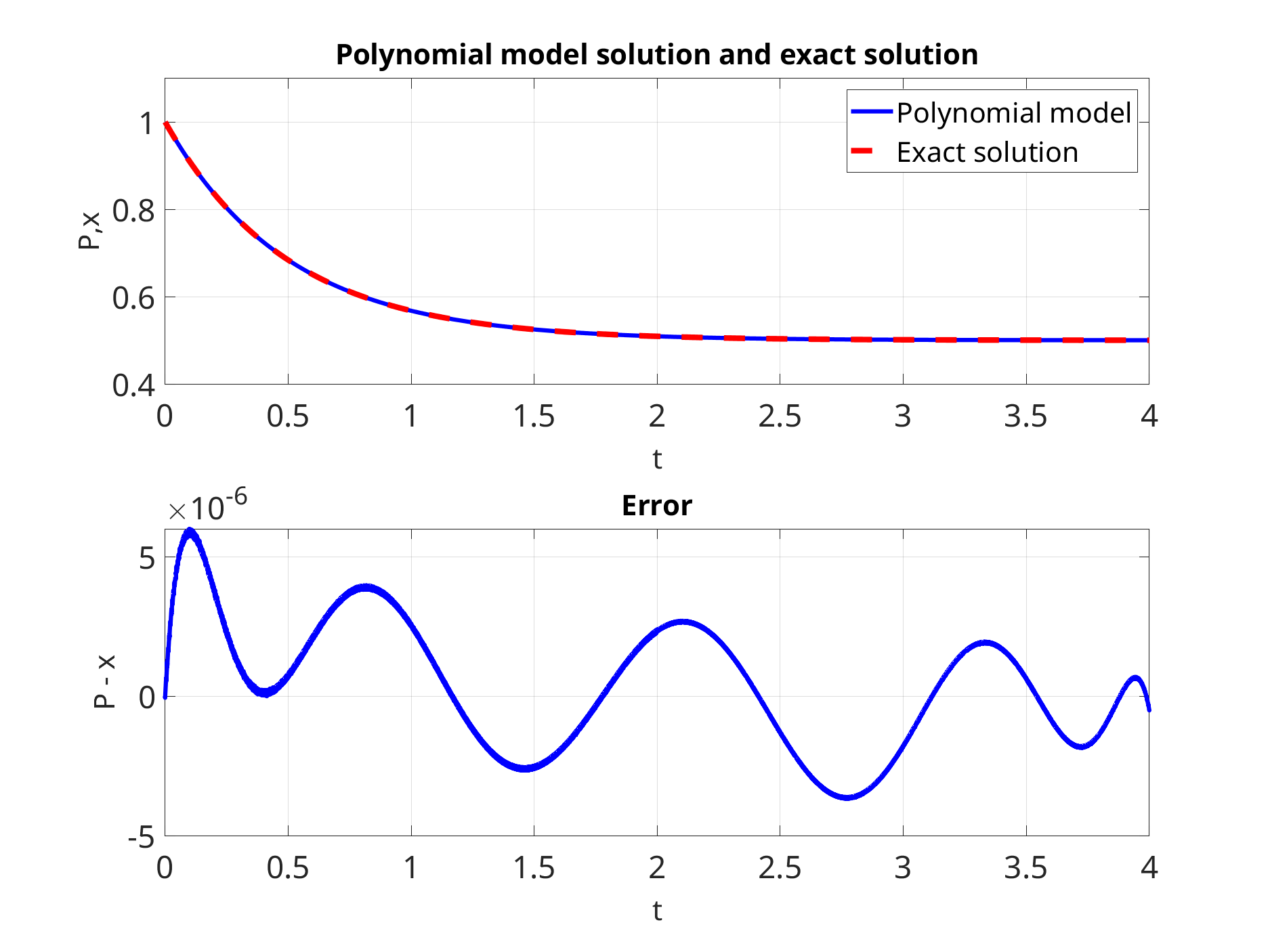}
        \caption{Type A: solution $x(t)$.}
        \label{fig:typeAf}
    \end{subfigure}\hfill
    \begin{subfigure}[t]{0.32\linewidth}
        \centering
        \includegraphics[width=\linewidth]{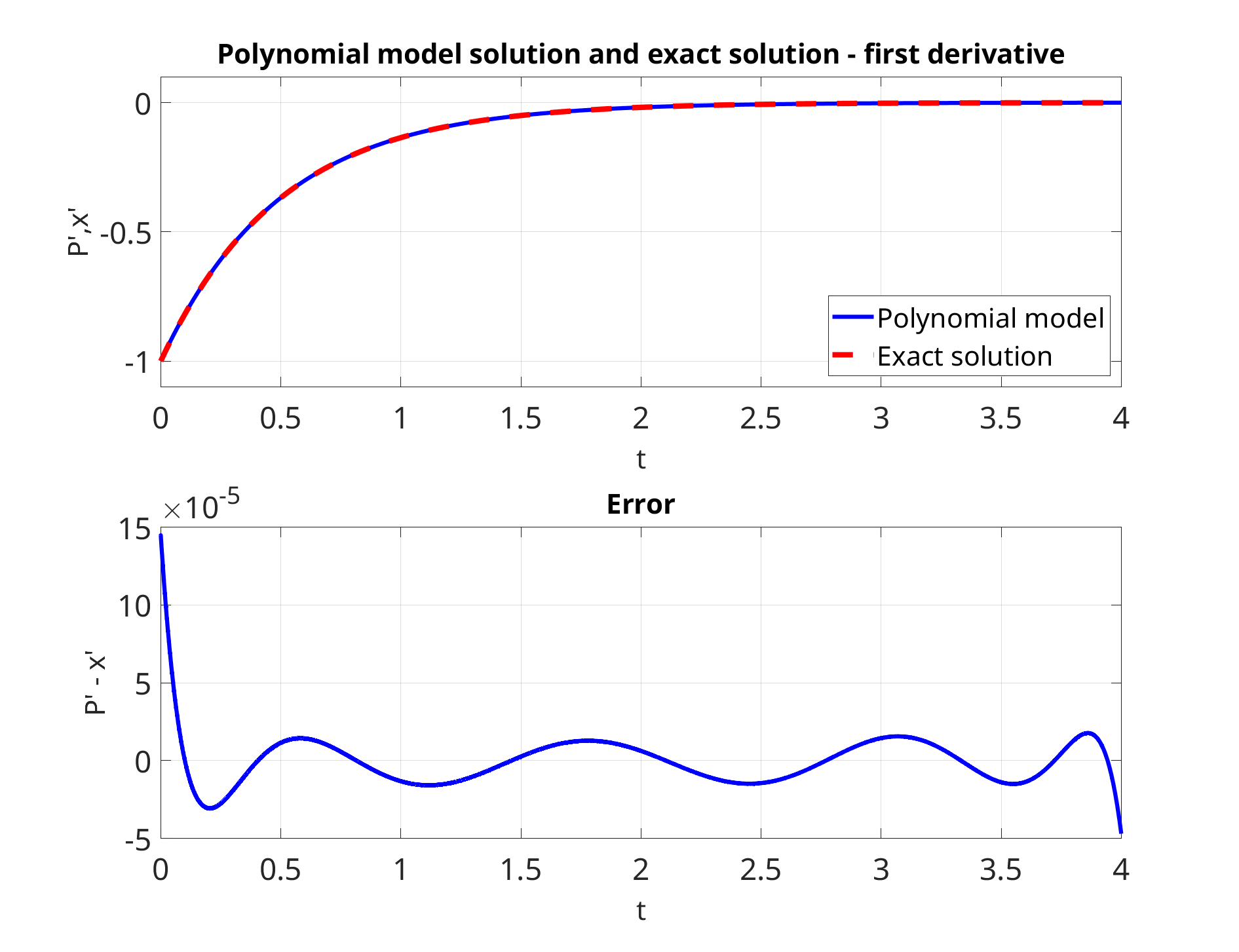}
        \caption{Type A: $x'(t)$.}
        \label{fig:typeAfd}
    \end{subfigure}\hfill
    \begin{subfigure}[t]{0.32\linewidth}
        \centering
        \includegraphics[width=\linewidth]{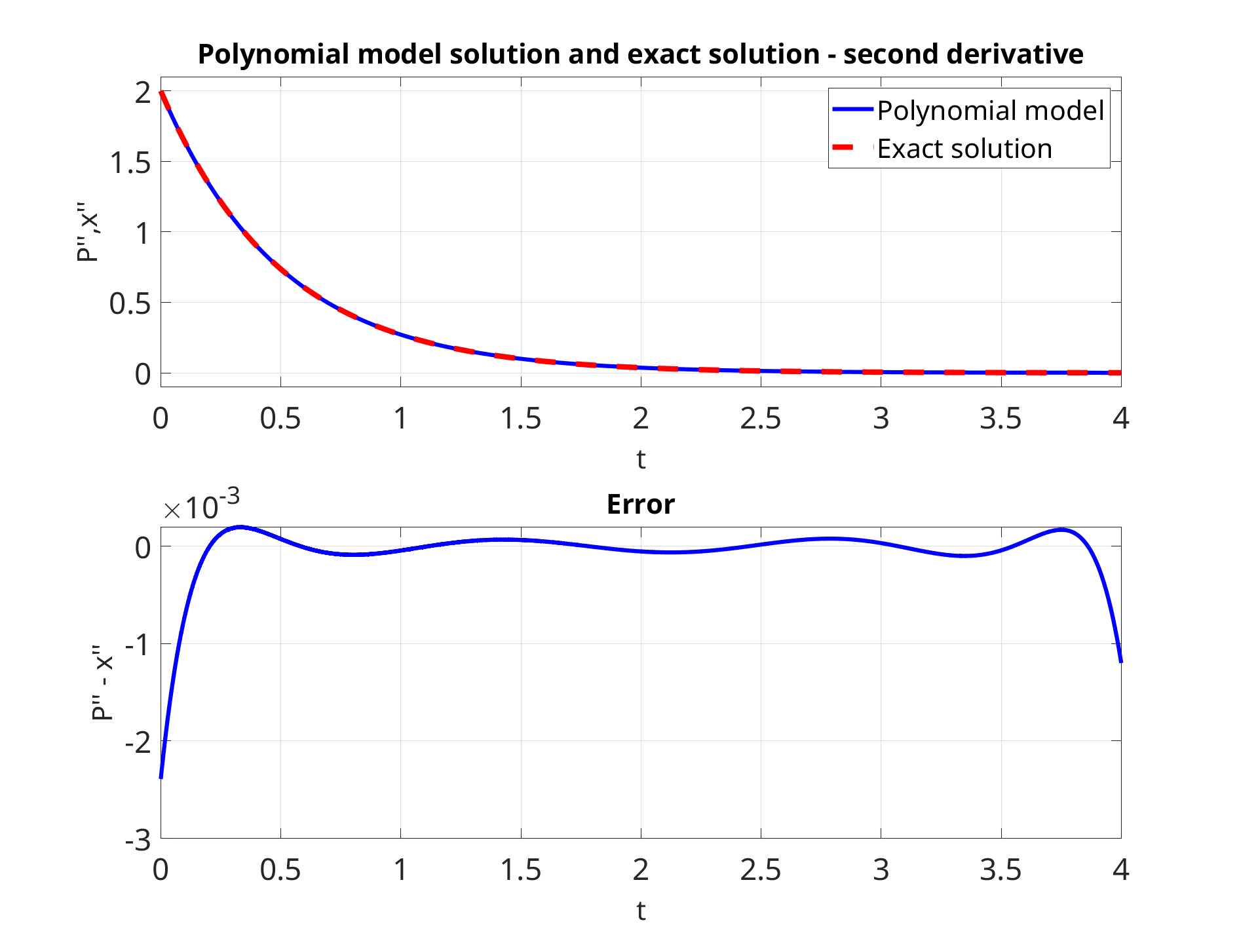}
        \caption{Type A: $x''(t)$.}
        \label{fig:typeAfdd}
    \end{subfigure}

    \vspace{2mm}

    \begin{subfigure}[t]{0.32\linewidth}
        \centering
        \includegraphics[width=\linewidth]{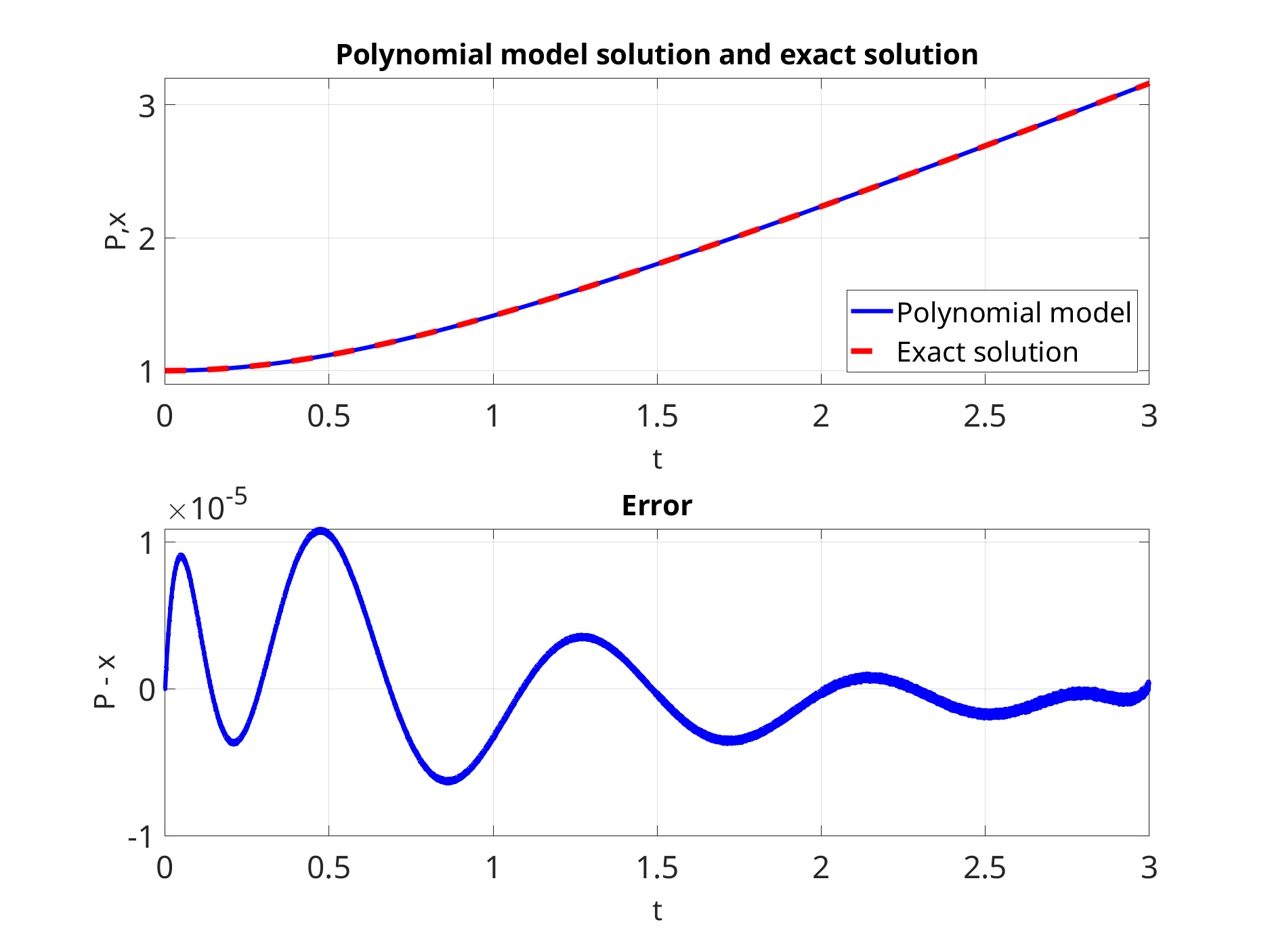}
        \caption{Type B: solution $x(t)$.}
        \label{fig:typeBf}
    \end{subfigure}\hfill
    \begin{subfigure}[t]{0.32\linewidth}
        \centering
        \includegraphics[width=\linewidth]{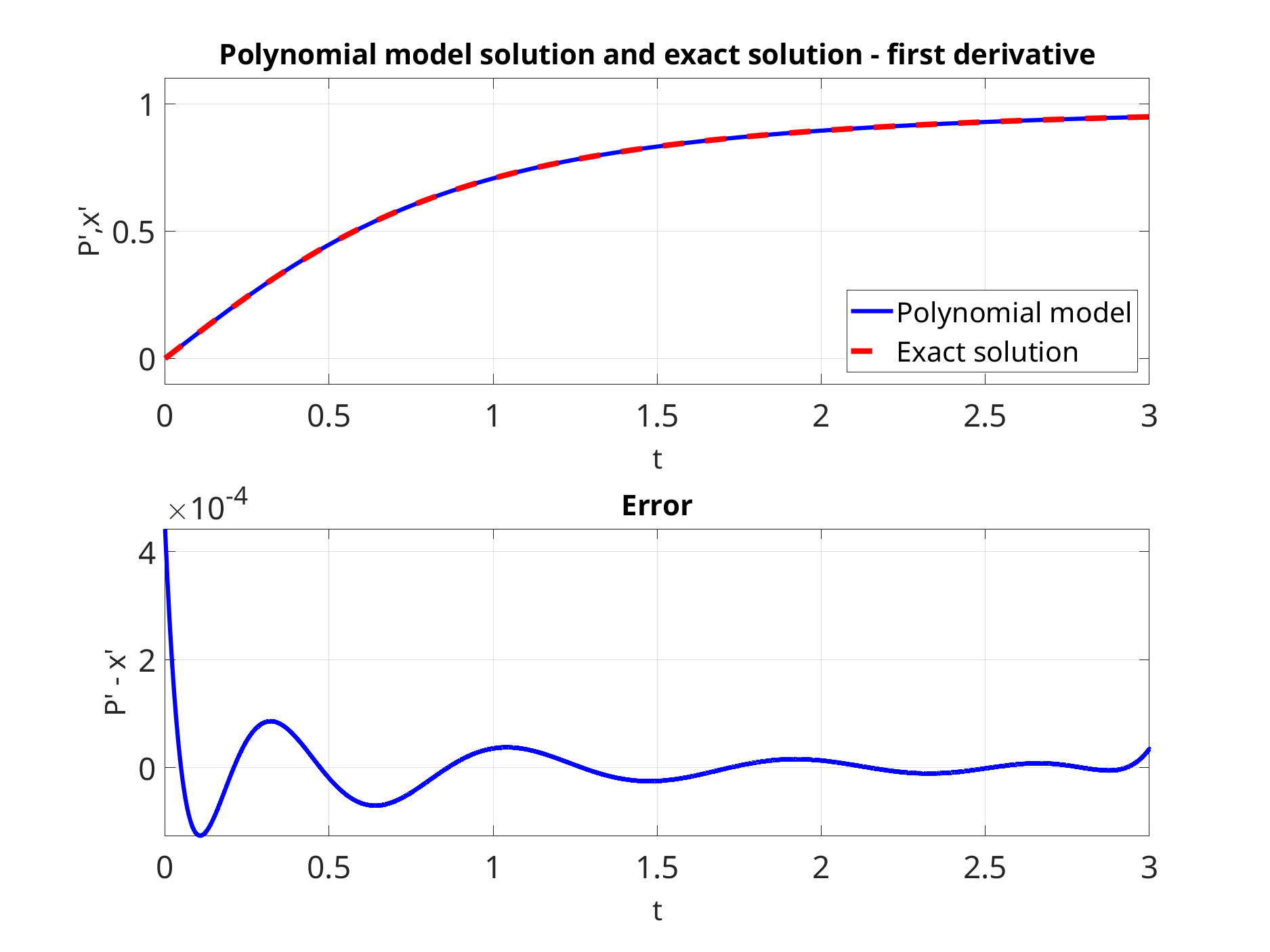}
        \caption{Type B: $x'(t)$.}
        \label{fig:typeBfd}
    \end{subfigure}\hfill
    \begin{subfigure}[t]{0.32\linewidth}
        \centering
        \includegraphics[width=\linewidth]{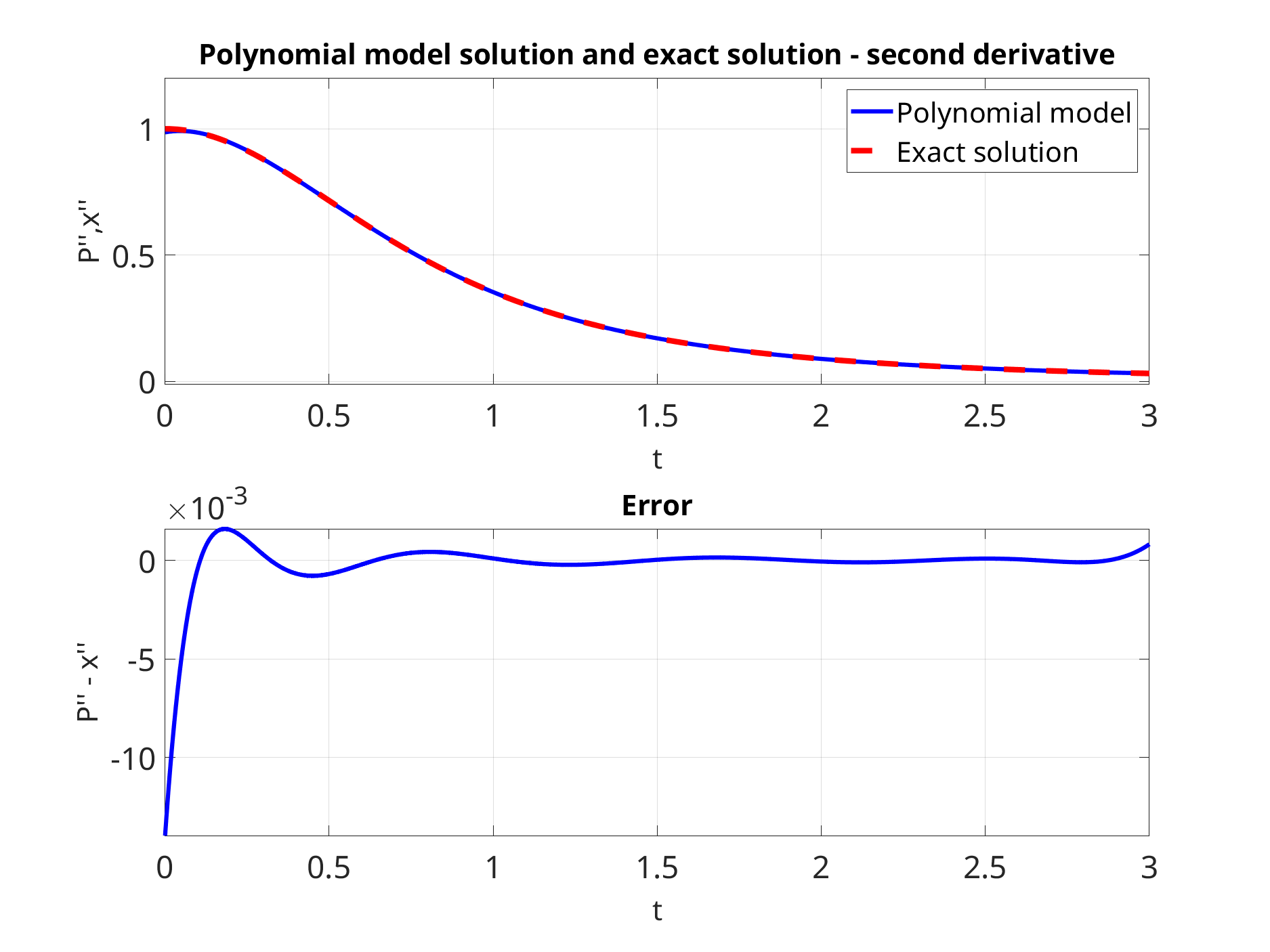}
        \caption{Type B: $x''(t)$.}
        \label{fig:typeBfdd}
    \end{subfigure}

    \vspace{2mm}

    \begin{subfigure}[t]{0.32\linewidth}
        \centering
        \includegraphics[width=\linewidth]{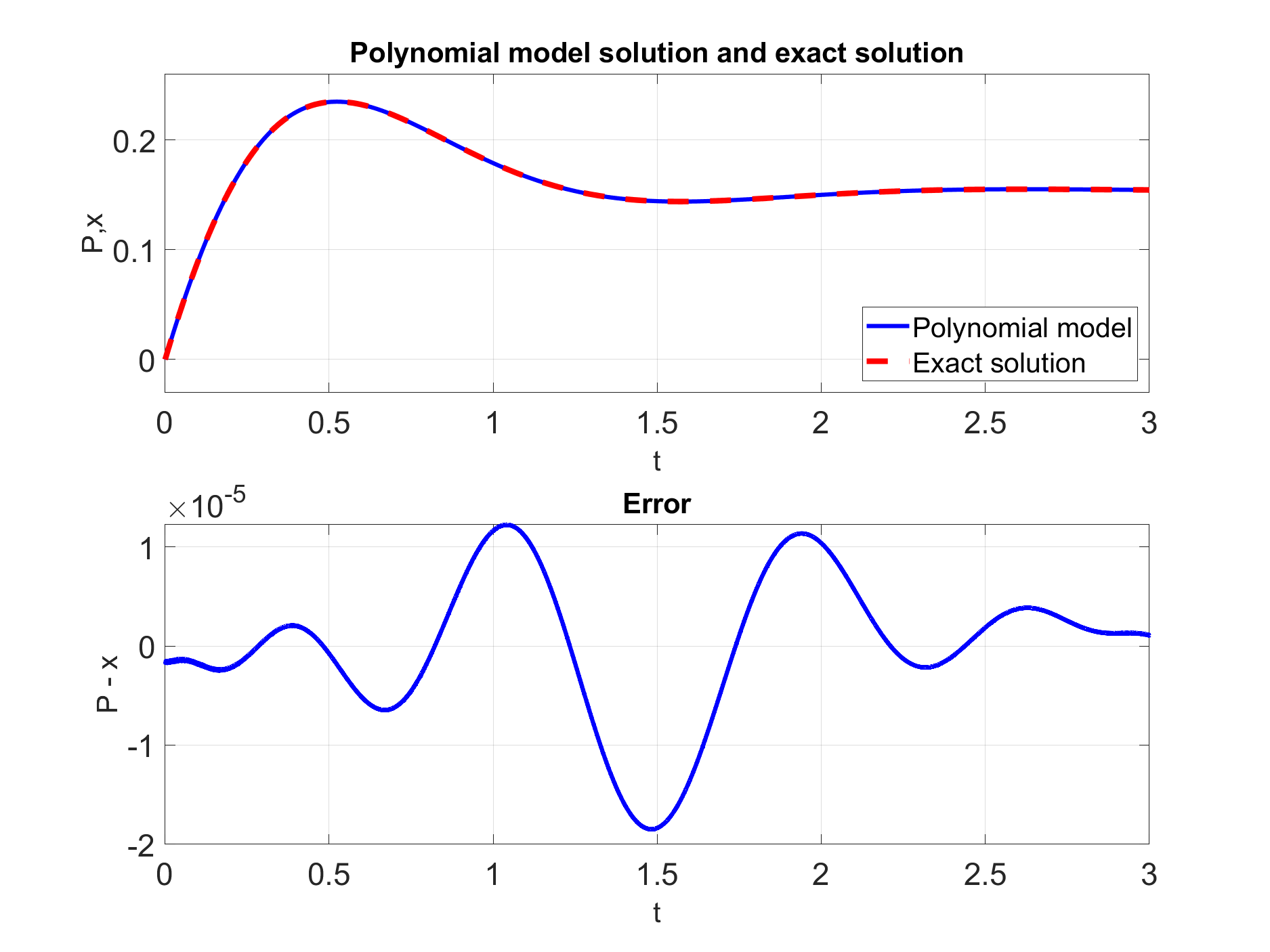}
        \caption{Type C: solution $x(t)$.}
        \label{fig:typeCf}
    \end{subfigure}\hfill
    \begin{subfigure}[t]{0.32\linewidth}
        \centering
        \includegraphics[width=\linewidth]{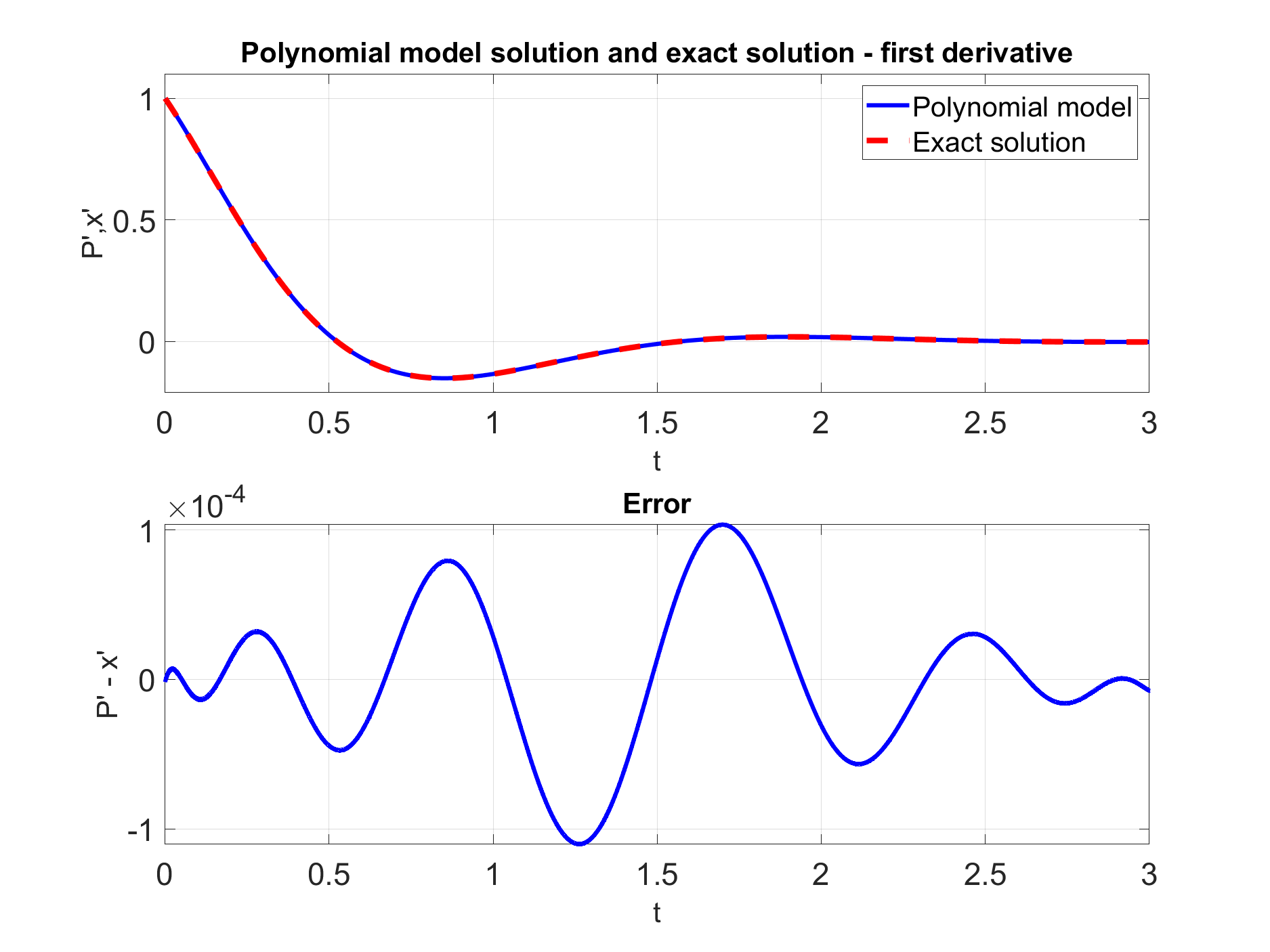}
        \caption{Type C: $x'(t)$.}
        \label{fig:typeCfd}
    \end{subfigure}\hfill
    \begin{subfigure}[t]{0.32\linewidth}
        \centering
        \includegraphics[width=\linewidth]{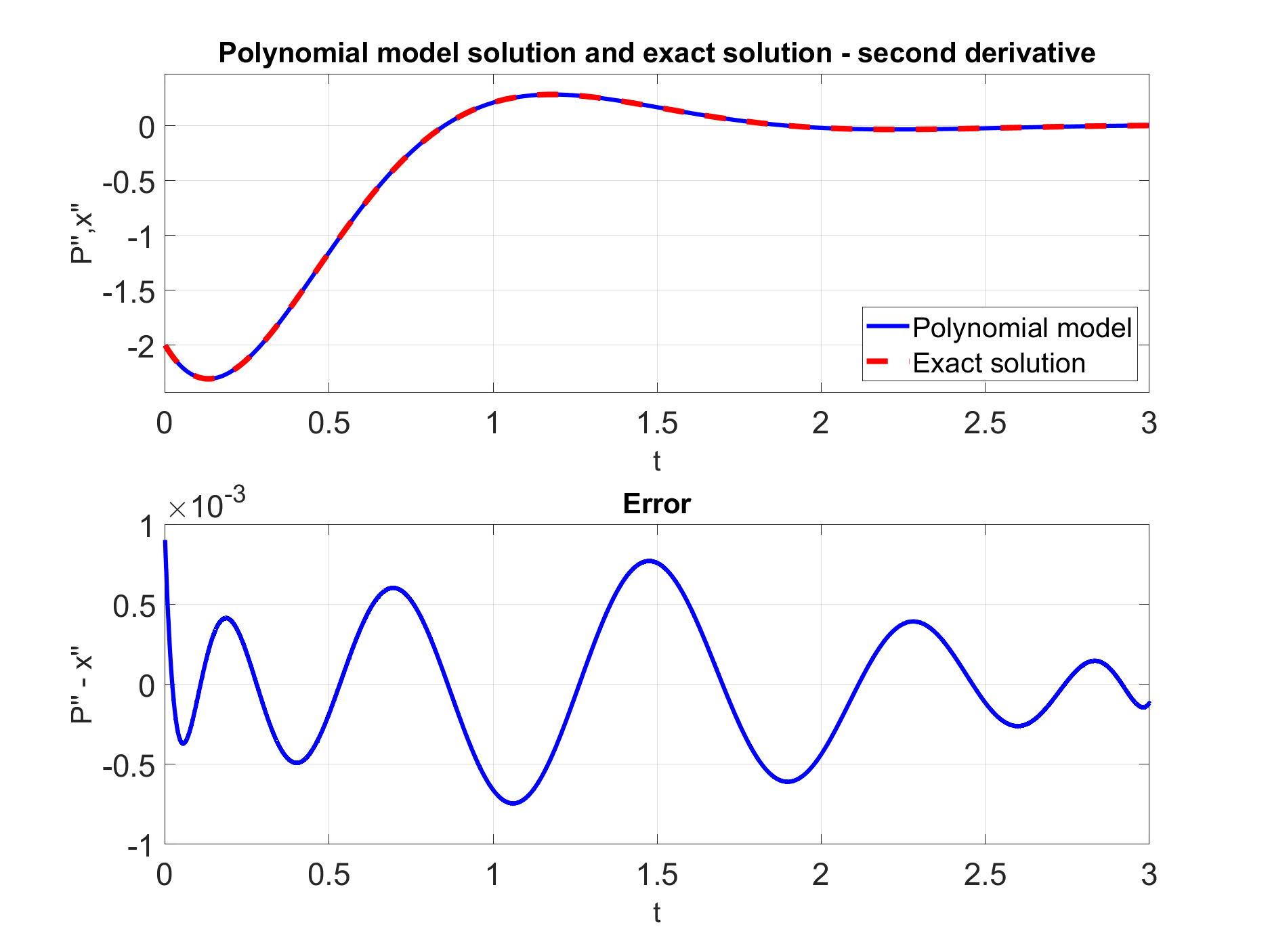}
        \caption{Type C: $x''(t)$.}
        \label{fig:typeCfdd}
    \end{subfigure}

    \caption{Horner network predictions for the three benchmark ODEs (Types A--C): solution and derivatives compared
    to analytical references.}
    \label{fig:Hornersol}
\end{figure}

For Types A and B we use a Horner network with 10 learnable parameters, while for Type C we use 13 learnable
parameters (the remaining low-order coefficients are fixed by the embedded initial conditions). For each benchmark,
we report the learned solution as well as its first and second derivatives (computed from the network representation).
Figure~\ref{fig:Hornersol} compares the Horner predictions to the analytical solutions.

Table~\ref{tab:res1} reports RMSE values for the solution as well as for the first and second derivatives,
computed on the same dense evaluation grid for all methods (lower is better; the best result in each row is shown in bold).
Across all three benchmarks, the proposed Horner network achieves the lowest RMSE while using markedly fewer learnable
parameters: 10 (Types A/B) or 13 (Type C) parameters, compared with $\mathcal{O}(10^2)$ parameters for the compact MLP (sigmoid)
and SIREN baselines and 263{,}937 parameters for the wide MLP with Leaky ReLU. This highlights a strong accuracy--efficiency
trade-off in favor of the proposed structured, parameter-minimal architecture.

\begin{table}[h]
\caption{RMSE on a dense uniform evaluation grid ($N_{\mathrm{eval}}=100\,000$ points on $I$) for the solution and its first two derivatives.
Lower is better; the best value in each row is shown in bold.}
\label{tab:res1}
\centering
\begin{tabular}{|c|c|c|c|c|c|}
\hline
 &  & LeakyReLU & Sigmoid & SIREN & Horner net \\ \hline
\multirow{3}{*}{Type A} 
 & $\mathrm{RMSE}(N,x)$ & $2.6\cdot10^{-3}$ & $8.5\cdot10^{-5}$  & $2.7\cdot10^{-5}$ & $\mathbf{4.6\cdot10^{-6}}$  \\ \cline{2-6}
 & $\mathrm{RMSE}(N',x')$ & $2.0\cdot10^{-2}$ & $5.6\cdot10^{-4}$  & $5.2\cdot10^{-4}$ & $\mathbf{3.3\cdot10^{-5}}$  \\ \cline{2-6}
 & $\mathrm{RMSE}(N'',x'')$& $1$ & $2.3\cdot10^{-2}$  & $2.7\cdot10^{-2}$ & $\mathbf{5.6\cdot10^{-4}}$  \\ \hline\hline
\multirow{3}{*}{Type B} 
 & $\mathrm{RMSE}(N,x)$   & $2.5\cdot10^{-3}$ & $2.7\cdot10^{-4}$ & $6.9\cdot10^{-4}$ & $\mathbf{6.6\cdot10^{-6}}$ \\ \cline{2-6}
 & $\mathrm{RMSE}(N',x')$  & $1.8\cdot10^{-2}$ & $1.5\cdot10^{-3}$ & $1\cdot10^{-3}$ & $\mathbf{7.8\cdot10^{-5}}$ \\ \cline{2-6}
 & $\mathrm{RMSE}(N'',x'')$ & $7.7\cdot10^{-1}$ & $1.4\cdot10^{-2}$ & $1.6\cdot10^{-2}$ &$\mathbf{2.3\cdot10^{-3}}$ \\ \hline\hline
\multirow{3}{*}{Type C} 
 & $\mathrm{RMSE}(N,x)$   & $1.3\cdot10^{-1}$ & $4.2\cdot10^{-4}$ & $1.1\cdot10^{-5}$ & $\mathbf{8.0\cdot10^{-6}}$ \\ \cline{2-6}
 & $\mathrm{RMSE}(N',x')$  & $1.7\cdot10^{-1}$ & $1.6\cdot10^{-3}$ & $9.1\cdot10^{-5}$ & $\mathbf{5.8\cdot10^{-5}}$ \\ \cline{2-6}
 & $\mathrm{RMSE}(N'',x'')$ & $1.5$ & $6.0\cdot10^{-3}$ & $1.3\cdot10^{-3}$ & $\mathbf{5.6\cdot10^{-4}}$ \\ \hline
\end{tabular}
\end{table}

\subsection{PDE - Horner Network}

In addition to ordinary differential equations that describe phenomena dependent on a single variable (e.g., time, position, temperature), many natural processes are influenced by multiple variables simultaneously. Such phenomena are modeled by partial differential equations (PDEs). 

We first explain how to generalize the 1D Horner network model presented in Section V.A to handle two variables. A general polynomial of order $n$ in two variables can be expressed as
\[
\begin{gathered}
P_2(x,y) = P_{1,n}(x) + y \cdot P_{1,n-1}(x) + y^2 \cdot P_{1,n-2}(x) + \dots + y^n \cdot P_{1,0}(x) = \sum_{i=0}^{n} y^{i} P_{1,n-i}(x),
\end{gathered}
\]
where $P_{1,k}(x)$ are polynomials of order $k$ in a single variable. This expression can be rewritten using Horner’s scheme as
\[
\begin{gathered}
P_2(x,y) = P_{1,n}(x) + y \left( P_{1,n-1}(x) + y \big( P_{1,n-2}(x) + \dots + y \big( P_{1,1}(x) + y \cdot P_{1,0}(x) \big) ... \big) \right).
\end{gathered}
\]
Each polynomial $P_{1,k}(x)$ can be implemented using the Horner network architecture introduced in Section V.A, denoted as $\mathcal{H}(P_{1,k})$. Moreover, the same architectural design generalizes to higher dimensions. The learnable parameters $a_0$, $a_1$, $\dots$, $a_m$ from Fig.~\ref{fig:BHBiHN} and Eq.~\ref{eq:Horner} are now entire neural networks $\mathcal{H}(P_{1,k})$.

We implemented this extended model to solve the initial-boundary value problem for the heat equation
\begin{equation}
\begin{gathered}
\frac{\partial u}{\partial t} - k \frac{\partial^2 u}{\partial x^2} = 0, \\
u(x, 0) = f(x), \\
u(0, t) = u(L, t) = 0,
\end{gathered}
\end{equation}
where we set $k = 0.1$, $L = 1$, and $f(x) = \sin(\pi x)$. The exact solution to this PDE is
\[
u(x,t) = \sin(\pi x) e^{-0.1 \pi^2 t}.
\]
The input data consists of a point cloud $\left(x_i, t_i, g(x_i, t_i)\right)$ for $i = 1, 2, \dots, M_1$, where $g(x, t)$ represents the excitation. The initial and boundary conditions are also provided as point clouds
\[
(x_j, 0, f(x_j)), \quad j = 1, 2, \dots, M_2,
\]
\[
(0, t_k, h_1(t_k)), \quad k = 1, 2, \dots, M_3,
\]
\[
(L, t_p, h_2(t_p)), \quad p = 1, 2, \dots, M_4.
\]
In our example, we used $M_1 = 5000$ and $M_2 = M_3 = M_4 = 2500$. The right-hand side of the PDE is set to zero ($g(x_i, t_i) = 0$) because the equation is homogeneous. The initial condition is $f(x_j) = \sin(\pi x_j)$, and the boundary conditions are $h_1(t_k) = 0$ and $h_2(t_p) = 0$. The initial and boundary conditions are enforced directly through the loss function
\[
\begin{gathered}
\mathcal{L}_{\text{loss}} = \frac{1}{M_1} \sum_{i=1}^{M_1} \left( F(t_i, \mathcal{N}(t_i) , \frac{d}{dt}\mathcal{N}(t_i), ..., \frac{d^n}{dt^n}\mathcal{N}(t_i)) \right)^2 + \frac{\lambda}{M_2} \sum_{j=1}^{M_2} \left( \mathcal{N}(x_j, 0) - f(x_j) \right)^2 + \\
\frac{\mu}{M_3} \sum_{k=1}^{M_3} \left( \mathcal{N}(0, t_k) - h_1(t_k) \right)^2 + 
\frac{\nu}{M_4} \sum_{p=1}^{M_4} \left( \mathcal{N}(L, t_p) - h_2(t_p) \right)^2.
\end{gathered}
\]
We set the hyperparameters as $\lambda = 0.5$ and $\mu = \nu = 0.25$. The total number of learnable parameters in the model is 45. Training is performed as described previously.

\begin{figure}[h]
    \centering
    \begin{subfigure}{0.32\textwidth}
        \centering
        \includegraphics[scale=0.35]{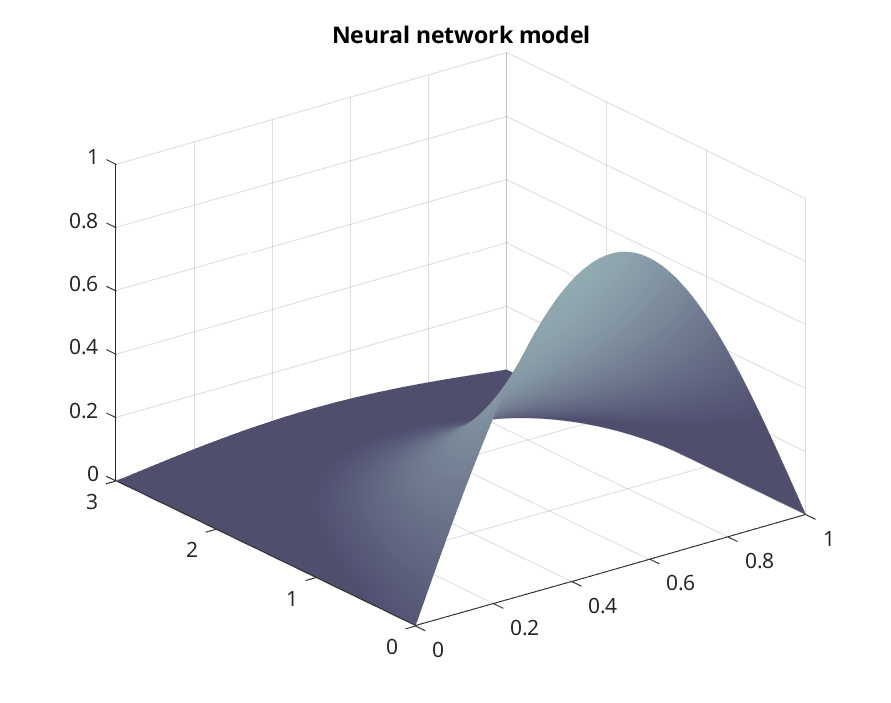}
        \caption{Horner model}
    \end{subfigure}
    \begin{subfigure}{0.32\textwidth}
        \centering
        \includegraphics[scale=0.35]{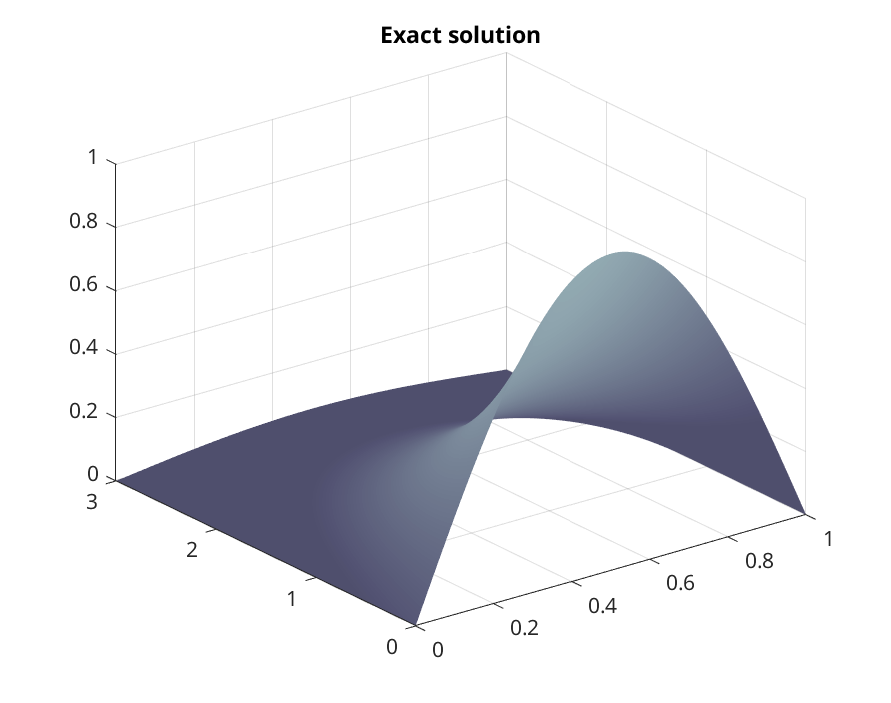}
        \caption{Exact solution}
    \end{subfigure}
    \begin{subfigure}{0.32\textwidth}
        \centering
        \includegraphics[scale=0.345]{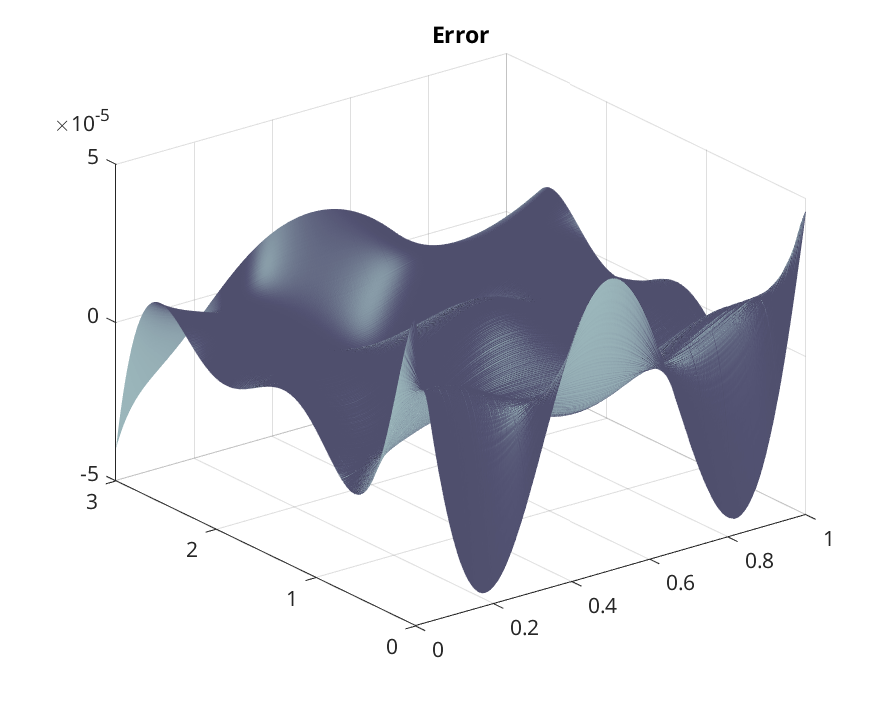}
        \caption{Error}
    \end{subfigure}
    \caption{Comparison of the solution obtained from the multidimensional Horner network with the exact solution of the heat equation. The error is minimal, even with only 45 learnable parameters.}
    \label{fig:heateq}
\end{figure}

Fig.~\ref{fig:heateq} compares the solution obtained from the neural network with the exact solution. The results demonstrate that the error is minimal despite using only 45 parameters, highlighting the efficiency and accuracy of the multidimensional Horner network. 

\section{Spline-like Model}

In this section, we extend the Horner network by introducing a spline-like approach to further reduce the approximation error when solving differential equations. The key advantage of this approach lies in its ability to balance accuracy with computational efficiency by keeping the number of learnable parameters to a minimum (slightly higher than the baseline model), even as the model complexity increases.

This technique is particularly beneficial for problems involving complex differential equations, where traditional single-network architectures may struggle to provide accurate solutions without a significant increase in the number of parameters. By leveraging the spline-like approach, we achieve an efficient trade-off between model size and accuracy, making it suitable for applications where computational resources are limited or precision is critical.

Let the interval of interest, where we solve the differential equation, be denoted as $[c, d]$. We partition this interval into subintervals such that $c = c_0 < c_1 < c_2 < \dots < c_l = d$. On each subinterval $C_i = [c_i, c_{i+1}]$, for $i = 0, 1, \dots, l-1$, we train a separate Horner network with a small number of parameters (corresponding to a lower-order polynomial). By employing a spline-like approach, we ensure through the loss function that the overall solution model belongs to the class $C^j([c, d])$ for a suitable value of $j$, ensuring continuity and smoothness.

\begin{figure}[h]
    \centering
    \includegraphics[scale=0.6]{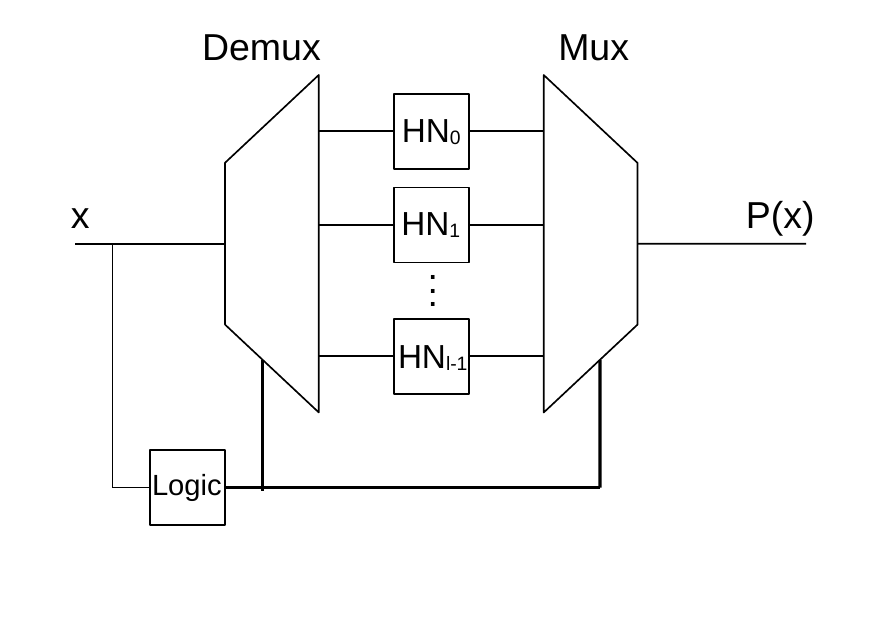}
    \caption{Spline-like model architecture. The input value $x$ determines which Horner network $HN_i$ is selected based on the interval it falls into, ensuring a piecewise smooth representation of the solution.}
    \label{fig:Splinelike}
\end{figure}

This procedure is illustrated schematically in Fig.~\ref{fig:Splinelike}. The input value $x$ is fed into a demultiplexer and logic module, which selects the corresponding network $\text{HN}_i$ based on the subinterval $C_i$ containing $x$. The output from the selected network is then passed through a multiplexer to produce the output of the entire spline-like model. Thus, we have $l$ Horner networks in total, with each network $\text{HN}_i$ responsible for learning the solution on its respective subinterval $C_i$. As stated earlier, the loss function enforces the model to be of class $C^j([c, d])$.

We test this model on a Type \textbf{a} differential equation. In this case, the loss function is defined as

\[
\begin{gathered}
\mathcal{L}_{\text{loss}} = \frac{1}{M} \sum_{i=1}^{M} \left( F(t_i, \mathcal{N}(t_i) , \frac{d}{dt}\mathcal{N}(t_i), ..., \frac{d^n}{dt^n}\mathcal{N}(t_i))  - u(t_i) \right)^2 +  \lambda_0 \left| \mathcal{N}(0) - x(0) \right| + \\ 
\sum_{j=0}^{l-2} \mu_j \left| \mathcal{N}_j(t_{j,r}) - \mathcal{N}_{j+1}(t_{j+1,l}) \right| +  
\sum_{j=0}^{l-2} \nu_j \left| \frac{d}{dt}\mathcal{N}_j(t_{j,r}) - \frac{d}{dt}\mathcal{N}_{j+1}(t_{j+1,l}) \right|. 
\end{gathered}
\]

Here, $t_{j,l}$ and $t_{j,r}$ denote the left and right endpoints of subinterval $C_j$. The third and fourth terms in the loss function enforce continuity of the solution and its first derivative at the subinterval boundaries.

For training the Type \textbf{a} differential equation solution, we used $M = 200$ data points. The hyperparameters were set to $\lambda_0 = 1$, $\mu_j = 0.5$, and $\nu_j = 0.5$. The subintervals were chosen as $[0, 1]$, $[1, 2]$, $[2, 3]$, and $[3, 4]$, resulting in a total of 4 Horner networks. Each network is characterized by 8 learnable parameters. We employ the Adam optimizer and the starting learning rate is set to $10^{-3}$. The network is trained for 10,000 epochs.

\begin{figure}[h]
    \centering
    \begin{subfigure}{0.32\textwidth}
        \centering
        \includegraphics[scale=0.2]{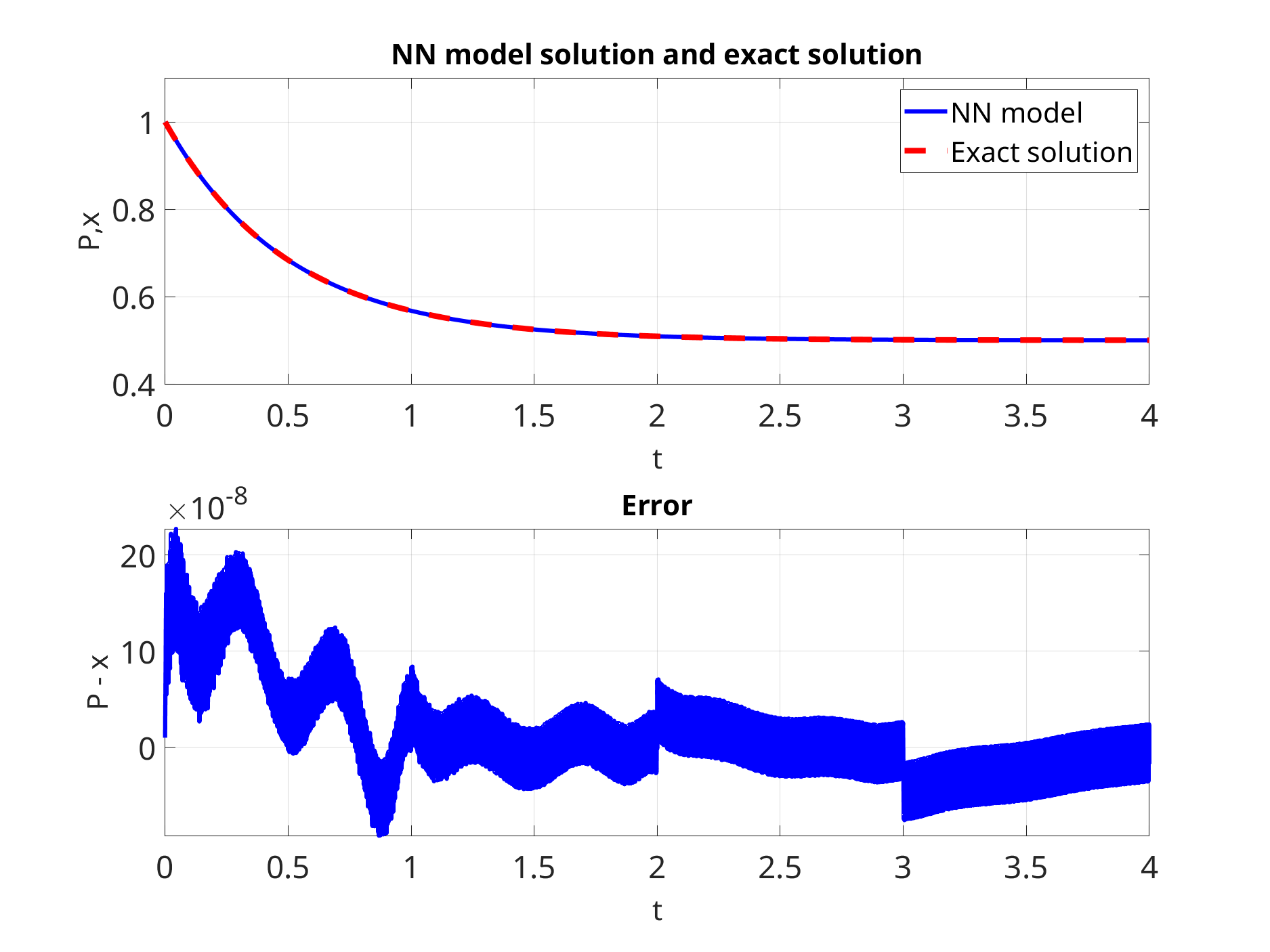}
        \caption{Type 1 - solution}
        \label{fig:s1}
    \end{subfigure}
    \begin{subfigure}{0.32\textwidth}
        \centering
        \includegraphics[scale=0.2]{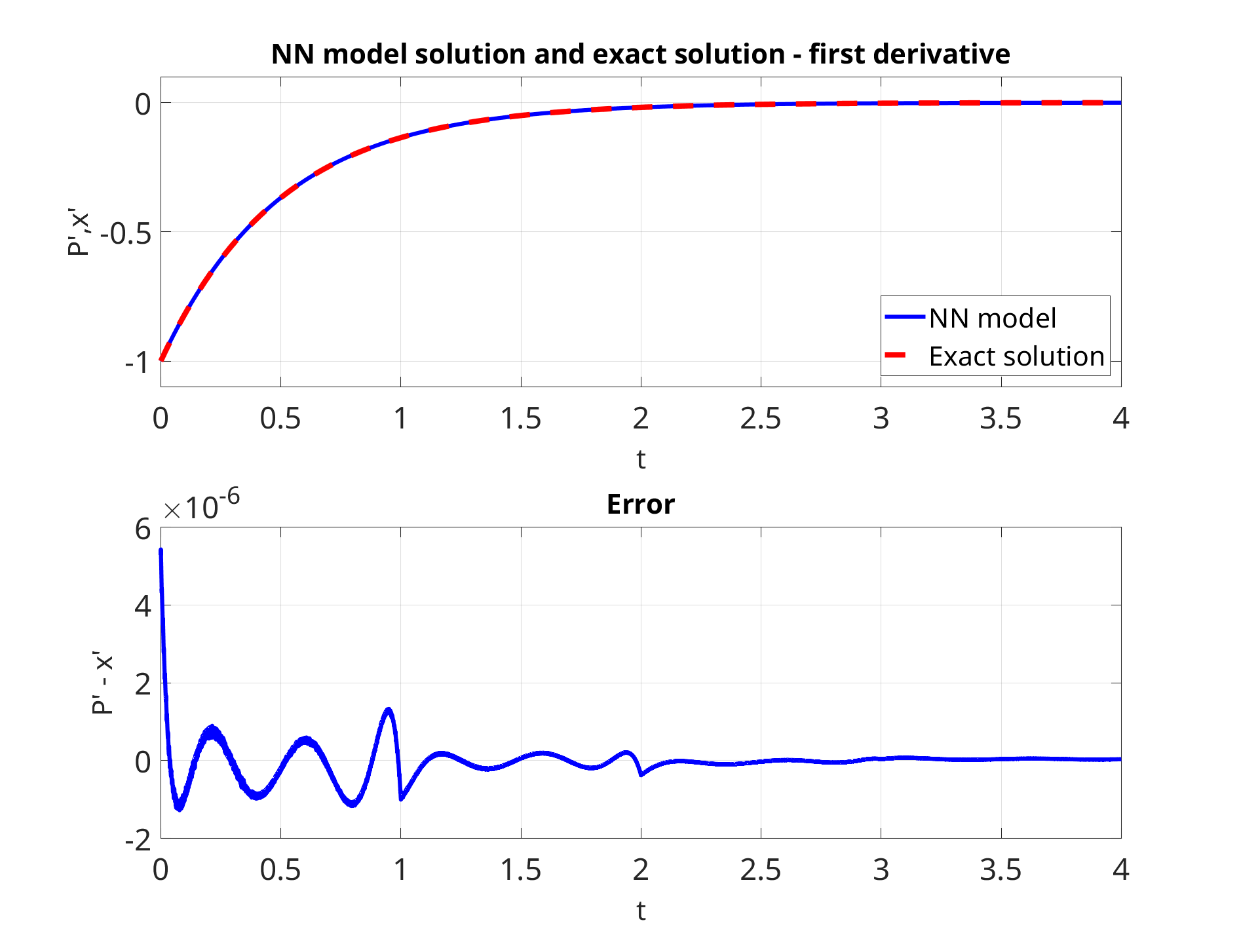}
        \caption{Type 1 - first derivative}
        \label{fig:s2}
    \end{subfigure}
    \begin{subfigure}{0.32\textwidth}
        \centering
        \includegraphics[scale=0.2]{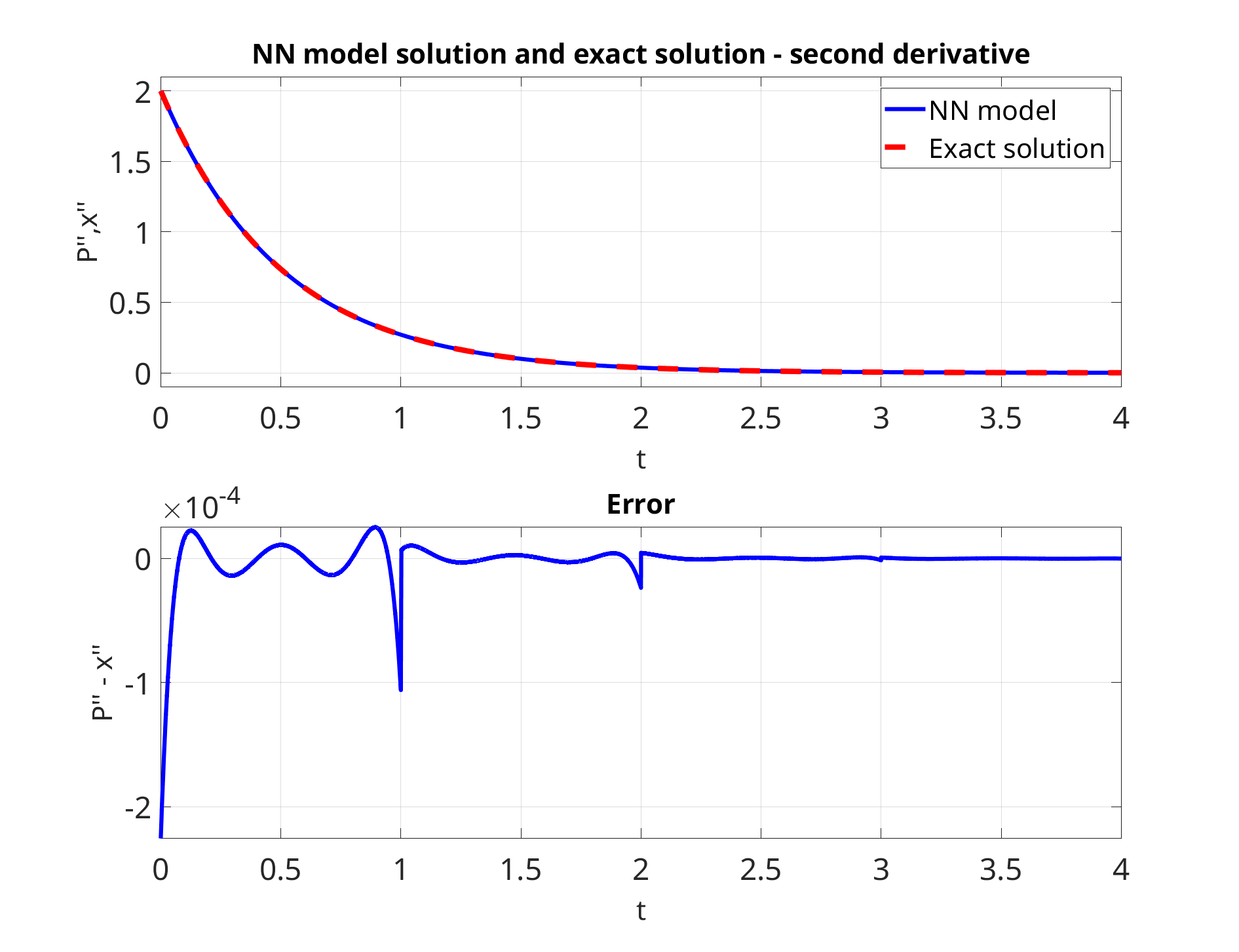}
        \caption{Type 1 - second derivative}
        \label{fig:s3}
    \end{subfigure}
    \caption{Comparison of the solution and its derivatives predicted by the spline-like Horner network with the exact solutions. The results demonstrate improved accuracy and smoothness across the subinterval boundaries.}
    \label{fig:splajnlajk}
\end{figure}

The results are presented in Fig.~\ref{fig:splajnlajk}. We observe that, in this example, the spline-like model achieves at least an order of magnitude improvement in accuracy over the baseline model presented in Section V.A.

\section{Conclusion}
In this paper, we have demonstrated that natural phenomena can be successfully modeled and solved using neural networks with a minimal number of learnable parameters. By leveraging carefully designed architectures, we have shown that even with a significantly reduced parameter count, accurate and reliable solutions to differential equations can be achieved.

Our model, based on the Horner scheme, achieved exceptional results with only around 10 learnable parameters. These models exhibited improvements of nearly three orders of magnitude in accuracy compared to standard MLP neural networks and networks with periodic activation functions, both of which typically require at least twice the number of parameters. This highlights the efficiency of our proposed architectures in scenarios where computational resources and parameter budgets are limited.

The spline-like approach further improved accuracy by effectively reducing the approximation error while only minimally increasing the number of learnable parameters. This method ensured continuity and smoothness across subinterval boundaries, making it particularly effective for applications requiring fine-grained precision and dynamic adaptation to problem complexity.

By solving the heat equation, we generalized the Horner-based approach to two-dimensional domains. Our results demonstrated that even with as few as 50 learnable parameters, we could accurately solve partial differential equations (PDEs). This shows that our method is not limited to one-dimensional problems but can be extended to more complex multidimensional scenarios with minimal computational overhead.

Further reductions in the number of learnable parameters in the spline-like model can be achieved by embedding boundary conditions of subintervals directly into the neural network architecture. This eliminates the need to enforce continuity or derivative constraints through the loss function, thus simplifying the training process and improving efficiency. We also propose potential extensions of the Horner-based approach to higher-order PDEs, offering exciting opportunities for addressing complex dynamic systems in future work.

An additional challenge lies in integrating the spline-like approach into models that solve complex partial differential equations (PDEs). This involves extending the current architecture to handle higher-dimensional domains and ensuring that continuity, smoothness, and physical boundary conditions are maintained directly through the network design. Overcoming this challenge will allow for scalable solutions to PDEs in fields such as fluid dynamics, electromagnetism, and 
thermodynamics, where traditional methods often face limitations in terms of parameter efficiency and computational complexity. A further challenge involves learning suitable subintervals for 1D cases, or patches in higher dimensions, to minimize the number of parameters or reduce the error, rather than keeping them fixed. Addressing these challenges will be a step in further advancing neural network-based approaches for scientific modeling.

In conclusion, our work provides a robust and scalable framework for solving differential equations using compact neural networks. Future efforts will focus on generalizing the approach to higher-order PDEs and exploring real-world applications in physics, engineering, and computational science, where accuracy, parameter efficiency, and adaptability are critical.


\bibliographystyle{IEEEtran}
\bibliography{ref.bib}
\balance

\end{document}